\documentclass[journal]{IEEEtran}
\usepackage{amsmath,amsfonts}
\usepackage{enumitem}
\usepackage{ragged2e}
\usepackage{algorithm}
\usepackage[caption=false,font=normalsize,labelfont=sf,textfont=sf]{subfig}
\usepackage{algpseudocode}
\usepackage{colortbl}
\usepackage{anyfontsize}
\usepackage{listings}
\usepackage{array}
\usepackage[table,xcdraw]{xcolor}
\usepackage{stfloats}
\usepackage{url}
\usepackage{verbatim}
\usepackage{graphicx}
\usepackage{booktabs}
\usepackage{bbding}
\usepackage{multirow}
\usepackage[accsupp]{axessibility}
\usepackage{pifont}
\usepackage[colorlinks=true,
            linkcolor=red,
            citecolor=blue,
            urlcolor=magenta,
            allbordercolors=white]{hyperref}
\usepackage{orcidlink}
\begin{document}

\title{Aperture-aware Dispersion 5-D Light-field Imaging Spectrometer}

\author{Chenglong Huang\orcidlink{ 0009-0007-4552-7737}, Tao Lv\orcidlink{0009-0006-8269-7623}, Jianing Yang\orcidlink{0009-0005-7232-7453},  Chongde Zi, Linsen Chen\orcidlink{0000-0002-1259-135X},  Xun Cao\orcidlink{0000-0003-3094-4371},~\IEEEmembership{Member, IEEE}
\thanks{Chenglong Huang, Tao Lv, Jianing Yang, Chongde Zi, Linsen Chen and Xun Cao are with Nanjing University, Nanjing, 210023, China. E-mail: $\lbrace$chenglong-huang, lvtao, jianing\_yang$\rbrace$ @smail.nju.edu.cn, $\lbrace$zichongde, chenls,  caoxun$\rbrace$ @nju.edu.cn.}
\thanks{Chenglong Huang, Tao Lv, Jianing Yang, Chongde Zi, Linsen Chen and Xun Cao are with the Key Laboratory of Optoelectronic Devices and Systems with Extreme Performances of MOE, Nanjing University.}
\thanks{Chenglong Huang and Tao Lv are co-first authors. Xun Cao is the corresponding author.}
}

\markboth{Journal of \LaTeX\ Class Files,~Vol.~14, No.~8, August~2021}%
{Shell \MakeLowercase{\textit{et al.}}: A Sample Article Using IEEEtran.cls for IEEE Journals}

\IEEEpubid{0000--0000/00\$00.00~\copyright~2021 IEEE}

\maketitle

\begin{abstract}
Enhancing perceptual dimensions while miniaturizing imaging systems presents significant challenges for high-dimensional visual sensing. Conventionally, the acquisition of the 5D $(x,y,u,v,\lambda)$ spectral light field (5D-SLF) data cube relies on bulky and expensive camera arrays, which are impractical for widespread application.
Existing single-detector systems are fundamentally limited by a trade-off between the resolutions of different dimensions owing to insufficient coding capabilities. Here we introduce an Aperture-aware Dispersion Light-field Imaging Spectrometer (ADLIS), that targets a synergy between compactness and resolution through aperture-multiplexed modulation, leveraging the inherent spectral-filtering properties of birefringent material.
Using only a manufacturing-friendly and cost-effective phase plate made of birefringent quartz crystal, the aperture of the proposed ADLIS enables compact angular-spectral encoding that is highly sensitive to both the incident angle and spectrum of incoming light.
In contrast to the viewpoint-separation approach of microlens arrays, ADLIS employs aperture encoding to superimpose all viewpoints onto each sensor pixel. This shifts the design paradigm from spatial division to encoding integration, aiming to achieve full-resolution light field recovery.
Thus, we develop the Aperture-aware Dispersion Light-field Imaging (ADLI) framework, which optimizes the aperture design and 5D-SLF reconstruction in an end-to-end (E2E) manner. Trained by simulation data and validated through real-world experiments, our system achieves robust high-performance 5D-SLF imaging while maintaining full spatial resolution.
\end{abstract}


\section{Introduction}
\IEEEPARstart{H}{igh-dimensional} light-field imaging is a pivotal frontier for advanced visual perception. Light-field imaging \cite{ziegler2017acquisition,stanford2008lightfield, ng2005light,wu2017light} overcomes the limited angular resolution of conventional planar imaging by capturing light rays from multiple viewpoints. Analogously, spectral imaging \cite{wagadarikar2008single,arce2013compressive,cao2011prism,cao2016computational,shi2025prior,deng2025compact,deng2025shared, chen2023notch} introduces perceptual capabilities across the spectral domain, empowering visual sensing technologies to probe intrinsic material properties of objects and thereby offering solutions for industrial inspection, material identification, and metamerism recognition. Furthermore, 5D-SLF imaging, which integrates mult-dimensional information to enable the acquisition of high-dimensional data, has facilitated groundbreaking advances in a wide array of scientific and application disciplines, such as autonomous driving \cite{gehrig2024low,almalioglu2022deep,zhao2017heterogeneous}, surgical navigation \cite{jakubovic2018high,kok2020accurate,matinfar2023sonification}, astronomic observation \cite{liodakis2022polarized, roberts2021rapid}, and environmental surveillance \cite{dai2023coastal}.

As a conventional solution, scanning-based systems reconstruct multi-dimensional light fields through dimensional scanning \cite{basedow1995hydice,hsu2017line,abdo2019spatial,stanford2008lightfield}. However, they are confined to static scenes, thus restricting their practical utility. Although multi-camera systems represent a typical approach for 5D-SLF acquisition \cite{zhu2018hyperspectral,holloway2014generalized,zhao2017heterogeneous}, their practical deployment is limited by bulky volume and complex calibration. A hybrid camera system \cite{xiong2017snapshot} that employs a Lytro branch along with a CASSI branch to encode angular and spectral information respectively, offers a viable alternative for 5D-SLF reconstruction. Unfortunately, this dual-branch system still requires complex calibration and precise alignment, which hinders its practical use outside laboratory environments.

In contrast,  encoding-based single-detector systems achieve high-dimensional acquisition in a more compact way.
Specifically, single-detector solutions capture 5D-SLF through modulating the incoming light into a 2D-sensor-available measurement, which contains rich spectral-angular information, and then recovering the original 5D cube computationally.
Representative systems \cite{zhu2018complete,marquez2019snapshot,lv2020snapshot,cui2020snapshot,marquez2020compressive, zhang2024compact} integrate a microlens array (MLA) with spectral encoders like color-coded aperture \cite{marquez2019snapshot,marquez2020compressive,zhang2024compact}, image mapping spectrometer (IMS) \cite{cui2020snapshot}, and birefringent polarization interferometer (BPI) \cite{zhu2018complete, lv2020snapshot}, to effectively encode the angular and spectral dimension respectively. However, the limitation of MLA-based solutions lies in their relatively low spatial resolution. It is because that each of the sensor pixel only records light ray from specific viewpoints instead of all, due to the secondary deflection generated by MLAs.
In fact, it is nontrivial to recover high-resolution high-dimensional light fields through an integrated and miniaturized imaging system. 
\IEEEpubidadjcol
Although some solutions acquire 5D-SLF by complicated optical encoders, such as catadioptric \cite{xue2017catadioptric}, metalens \cite{hua2022ultra} and rotated dove prism along with cylindrical lenses \cite{cui2021snapshot, zhao2023coded}, they significantly increase the cost of the system yet fail to ensure fabrication consistency, thus preventing broader adoption in practical applications. 

\begin{figure}[!tb]
\centering
\includegraphics[width=0.95\linewidth]{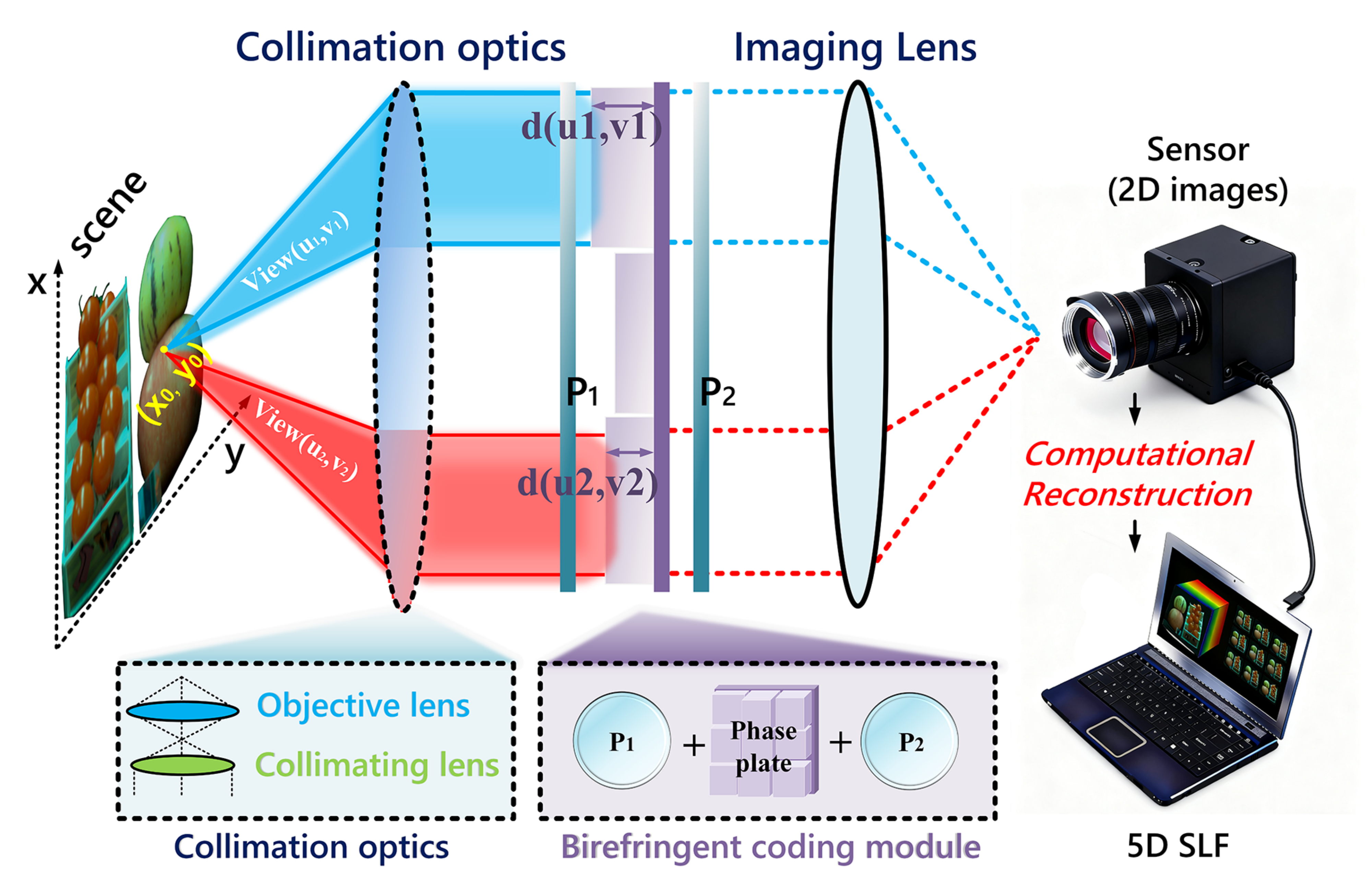}
  \caption{ Ray-tracing of the light propagation through the proposed ADLIS. Light from the scene sequentially passes through the collimation optics, the birefringent coding module, and the imaging lens, ultimately forming single- or multi-frame 2D measurements on the sensor (depending on whether P2 is rotated). These measurements are then used to reconstruct the 5D-SLF by the computational imaging framework.}
  \label{fig:optical}
\end{figure}

All in all, despite the fact that some methods mentioned above, to some extent, can compress 5D-SLF to a 2D measurement, making it real-timely available for 2D sensors, none of them can recover high-precision 5D-SLF at full resolution without sacrificing the spatial sampling rate owing to the dimension trade-off.
\begin{figure*}[!htb]
\centering
\includegraphics[width=1\linewidth]{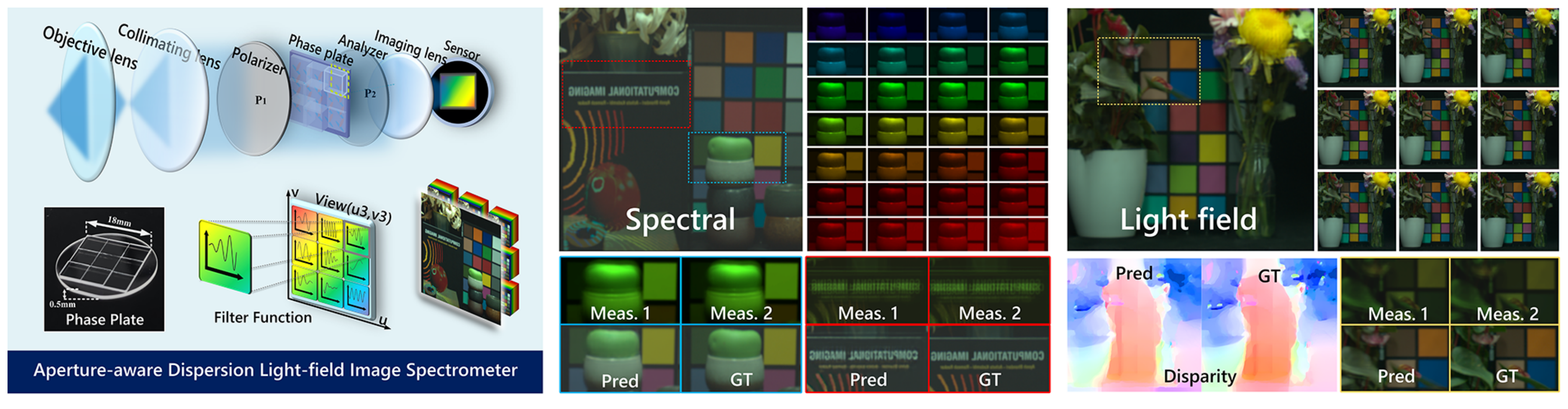}
  \caption{ (Top left) Schematic diagram of the aperture-aware dispersion light-field imaging spectrometer (ADLIS). Here, we present the physical photograph of the phase plate and a schematic diagram of its filtering function (bottom left). We also present the overall spectral performance (center) and angular performance (right) of the proposed ADLIS. Specifically, we detail the visualization of 2 measurement frames, prediction results and ground truth. The ADLIS system demonstrates positive outcomes in both spectral band-by-band analysis and light field disparity estimation. }
  \label{fig:one}
\end{figure*}
In this work, we demonstrate a novel single-detector 5D-SLF imaging system called the aperture-aware dispersion light-field imaging spectrometer (ADLIS) by developing a compact birefringent coding model (BCM), which is able to recover full-spatial resolution light field. 
The encoding process of the proposed ADLIS is conducted on the collimated aperture plane through a phase plate made of birefringent quartz crystal as is shown in Fig. \ref{fig:one} and Fig. \ref{fig:optical}. 
Given that the birefringent phase plate exhibits a spectral filtering response modulated by its thickness, we specifically designed it so that the aperture produces distinct spectral responses to incident light from different viewpoints. This design enables the integration of both light field sensing and spectral dispersion within a single, aperture-plane optical element.
What makes this approach significant is that ADLIS projects the modulated incoming light from all viewpoints on the complete sensor in a spatially multiplexed manner. Consequently, unlike traditional MLA-based paradigm, each sensor pixel no longer corresponds to some specific viewpoints but instead carries information integrated from all perspectives in ADLIS pipeline. This strategy obviously  increases the throughput of encoded information, thereby establishing the foundation for computationally reconstructing a full-resolution light field subsequently.
Moreover, the thickness parameter of the phase plate, defined herein as a thickness map, is fully differentiable. Accordingly, we propose an E2E ADLI framework that reconstructs the 5D-SLF from 2D measurements and simultaneously achieves an optimized optical design.
By leveraging high-quality data \cite{li2025realslf} as a physical constraint and a compact angular-spectral encoding module, ADLIS achieves 5D-SLF imaging without trading spatial resolution for angular acquisition, thereby fully preserving spatial information throughout the image formation process. Through extensive numerical simulations, we identify the optimized optical design and finally build an ADLIS prototype, whose superior performance is further validated through real-world experiments.
To summarize, our main contributions can be listed as follows:

\begin{itemize}
    \item We propose a full-resolution encoding model tailed for high-dimensional spectral light-field imaging, which is compact and effective, and present a co-optimized framework to recover the full-resolution 5D-SLF.
    \item Our approach involves an angular-spectral-aware phase plate made of birefringent quartz crystal on the aperture plane, which is affordable and thickness-differentiable.
    \item The proposed E2E framework outperforms conventional color-coded methods by a large margin and demonstrates notable robustness to variations in network architecture of the decoder.
    \item We develop the ADLIS prototype equipped with custom phase plates as its encoding elements. Our system is rigorously validated on both simulations and real-data experiments.
\end{itemize}

\section{Related Work}

\subsection{Coded-aperture Computational Imaging, CCI}
To obtain information with more dimensions without introducing additional acquisition devices, CCI serves as an effective approach for a single 2D sensor to capture multidimensional light fields. In recent years, the rapid development of AI, especially deep neural networks, have advanced the design of E2E imaging systems. These systems utilize encoding elements to modulate the amplitude, phase, and polarization of incident light, and recover extra-dimensional information including spectral \cite{wagadarikar2008single,cao2011prism,chen2023notch}, angular \cite{ng2005light}, depth \cite{ghanekar2024passive}, time \cite{meng2020gap,qiao2020deep} and polarization \cite{fan2024dispersion,fu2025miniaturized} through computational reconstruction. Moreover, multidimensional imaging systems also benefit from the development of coded aperture design. Lopez et al. \cite{lopez2024depth} design a color coded aperture (CCA) with a greater number of color filters and richer spectral information to optically encode relevant depth information in a single snapshot. Lv et al. \cite{lv2020snapshot} investigate a snapshot spectral-polarimetric-volumetric imaging system through several optical encoding elements.
Paradigms based on CCI mentioned above demonstrates an increasing progress in miniaturization and real-time capabilities recently. However, it still poses a formidable challenge when stringent requirements are imposed on both the imaging dimensions and resolution of the system.

Our work aims to leverage a superior aperture-coded strategy, which is well-suited for high-dimensional light-field imaging, to enable a conventional 2D sensor to simultaneously acquire spectral and angular dimensions, thereby achieving 5D-SLF imaging. This design excels in its low cost, compact form factor, and high encoded information throughput, while further facilitating E2E optimized design leveraging its differentiable optical parameters.

\subsection{Spectral Light-field Imaging}
5D-SLF aims to acquire high-dimensional light field information in 2D space $(x,y)$, 2D angle $(u,v)$, and 1D spectrum $(\lambda)$. Capturing high-dimensional scene information with measurements from a 2D sensor remains a significant challenge. Previously, Xiong et al. achieved 5D light-field spectral imaging \cite{xiong2017snapshot} through a hybrid system that distributes the scene information to a light-field camera and a snapshot hyperspectral camera through a beam splitter, combined with a multiplexed image fusion method; Zhao and Zhu et al. \cite{zhu2018complete, zhao2017heterogeneous} also proposed the systems that use the camera array with band-pass filters to acquire SLF. However, all of the aforementioned system configurations require multiple cameras, resulting in complex joint calibration and an exponential increase in system cost. MLA-based paradigms \cite{lv2020snapshot, marquez2020compressive, zhang2024compact} facilitate a certain degree of miniaturization for SLF imaging systems. However, this comes at the inevitable cost of spatial resolution, as they must allocate limited 2D sensor pixels to encode the angular information. 

Enabled by recent progress in neural networks, the synergy between data-driven reconstruction and compact encoding model has emerged as a powerful solution for multi-dimensional computational imaging \cite{lv2023aperture,shi2023compact,lv2024efficient, cai2024exploring}. We work on this path to propose an imaging system that records and recovers 5D-SLF in the full-spatial resolution.

\subsection{Deep Optics}
The idea of E2E joint optimization of optical encoder and computational decoder has rencently gained much attention. This approach has led to notable performance and improvements across various imaging applications. Sitzmann et al. \cite{sitzmann2018end} applied deep optics to extended depth of field and super-resolution tasks, achieving crucial enhancements in image quality. Chang et al. \cite{chang2019deep} utilized deep optics framework for monocular depth estimation and 3D object detection, while Ikoma et al. introduced an occlusion-aware image formation model that further improved accuracy in monocular depth estimation \cite{ikoma2021depth}.  Shi et al. \cite{shi2024split} achieves the simultaneous capture of coded and uncoded images in a single exposure by jointly optimizing the phase modulation and image reconstruction algorithms of diffractive optics through a differentiable wave optics model, which significantly improves the performance of high dynamic range, depth and spectral reconstruction. Additionally, E2E deep optics has shown strong performance in complex scenarios including hyperspectral imaging \cite{zhang2022herosnet, shi2018hscnn+, li2022quantization}, computational microscopy \cite{hershko2019multicolor, horstmeyer2017convolutional, nehme2020deepstorm3d}, time-of-flight imaging \cite{marco2017deeptof, su2018deep}, imaging through scattering media \cite{turpin2018light} and extended depth-of-field imaging \cite{yang2024end}.

In our work, we design an E2E framework that parameterizes the response function of the birefringent coding model as globally differentiable variables. It allows for joint optimization of the optical parameters and the high-dimensional light-field reconstruction network, achieving a better match between the optical design and the reconstruction network, facilitating the search for a globally optimal design. To the best of our knowledge, this work is the first to explore E2E deep optics framework for 5D-SLF imaging.

\section{Proposed Approach}
\subsection{Image Formation}\label{subsec:formation}
\begin{figure}[!htb]
\centering
\includegraphics[width=1\linewidth]{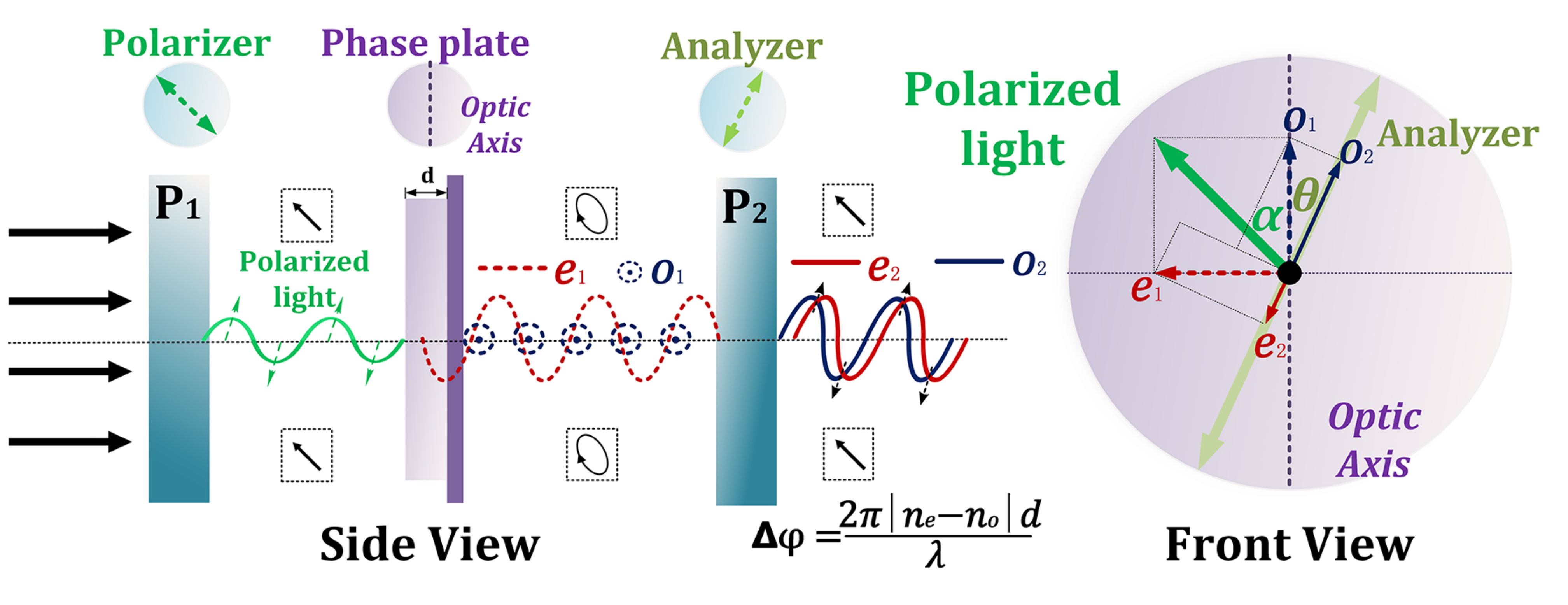}
  \caption{Schematic diagram of BCM when acting on linearly polarized light.}
  \label{fig:two}
\end{figure}

A schematic of the proposed ADLIS is shown in Fig.~\ref{fig:one}. The ADLIS consists of an objective lens, collimating lenses, and the birefringent coding module (BCM) comprising a polarizer, birefringent quartz crystal and a rotatable analyzer. The ADLIS operates as follows: incident light rays from different angles are collimated and then encoded by the BCM before being captured by the imaging lens to form measurements on the sensor, which is detailed in Fig. \ref{fig:optical}. By leveraging the BCM and optimizing the optical parameters, we achieve angle-dependent aperture encoding.
Specifically, ADLIS encodes angular information at the aperture plane in a way that allows the modulated results from each viewpoint to be multiplexed spatially onto every pixel of the image sensor. This enables the sensor to retain its full native spatial sampling rate, thereby avoiding the spatial-angular trade-off inherent to MLA-based light field systems.
In ADLIS, we use a birefringent material to encode the 5D-SLF. Birefringent crystals, such as quartz, exhibit anisotropic optical properties characterized by two different refractive indices: \(n_o\) and \(n_e\). As shown in Fig. \ref{fig:two}, in the BCM, when linearly polarized light passes through a birefringent crystal, it is split into two orthogonal polarization components, \(I_o\) and \(I_e\), known as o-light and e-light, respectively. This process introduces a phase difference between the two components, converting linearly polarized light into elliptically polarized light.
Mathematically, the phase difference induced by \(n_o\) and \(n_e\) is expressed as
\begin{equation}
\Delta \varphi = \frac{2 \pi d \Delta n}{\lambda} = \frac{2 \pi}{\lambda} \left|n_e - n_o\right|d,
\label{2-1-1}
\end{equation}
where $d$ is the thickness of the birefringent quartz crystal and $\lambda$ is the wavelength of the
incident light. The light intensity behind the analyzer is
\begin{align}
I &= {I_0}\cos^{2}(\alpha-\theta)-{I_0}\sin2\alpha\sin2\theta\sin^{2}(\frac{\Delta \varphi}{2})
\notag \\
&= {I_0}\cos^{2}(\alpha-\theta)-{I_0}\sin2\alpha\sin2\theta\sin^{2}(\frac{\pi \left|n_e - n_o\right|d}{\lambda}),
\label{2-1-2}
\end{align}

\begin{figure*}[!htb]
\centering
\includegraphics[width=1\linewidth]{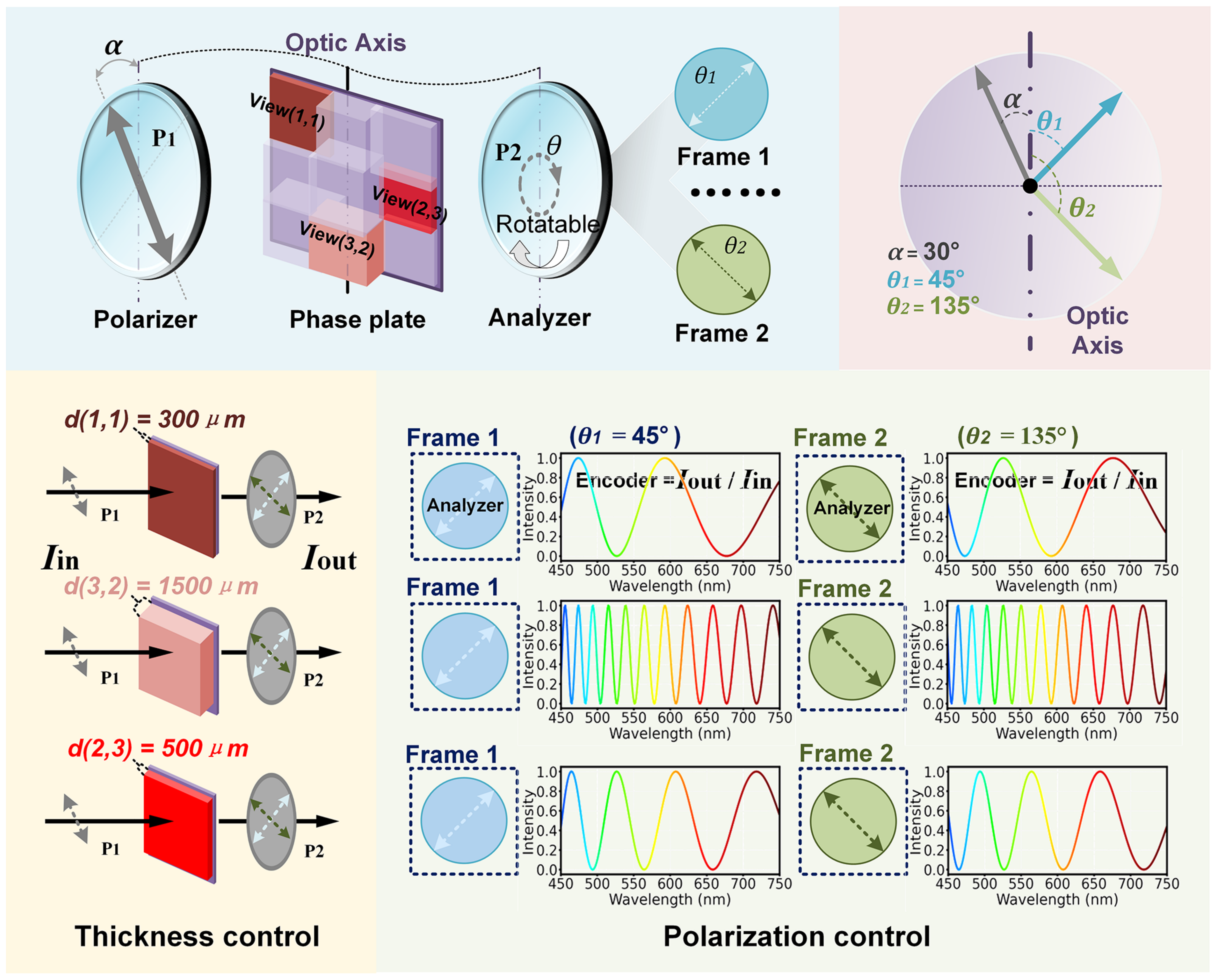}
  \caption{ (Top left) Overall of the BCM. The encoding function of the aperture is jointly determined by the thickness map of the birefringent quartz phase plate and the orientation of the analyzer. (Top right) Front view. The angle between the polarizer and the optical axis is denoted as $\alpha$, while $\theta_1$ and $\theta_2$ represent the analyzer orientations corresponding to two distinct measurement frames in the encoding scheme. (Bottom) Thickness control and polarization control. The spectral period is set by the thickness, and the key to optimal encoding is selecting analyzer angles that yield opposite spectral responses for the two frames.}
  \label{fig:pola}
\end{figure*}

where ${I_0}$ denotes the initial incident light intensity, $\alpha $ and $\theta $ are the angles between the optical axis and the polarizer/analyzer, respectively.
Evidently, the filter function of this system with respect to the wavelength is
\begin{equation}
H=\frac{I}{I_0}=\cos^{2}(\alpha-\theta)-\sin2\alpha\sin2\theta\sin^{2}(\frac{\pi\left|n_e - n_o\right|d}{\lambda}).
\label{2-1-3}
\end{equation}
For a given combination of ($n_e$, $n_o$, $\alpha$) = (1.5440, 1.5519, 30°), the filter function of the system $H(d,\theta)$ can be directly modulated by the thickness $d$ and the angle $\theta$. 
To achieve compact angle-dependent modulation, we fully leverage the tunable optical parameters $d$ and $\theta$ to design the phase plate as an aperture array. As shown in Fig. \ref{fig:optical}, light rays from the same scene point at different angles pass through distinct regions of the aperture plane, enabling the designed aperture array to perform spectral dispersion while simultaneously encoding the angular information.
Fig. \ref{fig:pola} presents the detailed principle of the BCM, which is controlled by a thickness-polarization dispersion mechanism.
The dispersion encoding function experienced by each light ray is directly governed by the thickness of the phase plate it traverses, a variation that depends on the incident angle of the ray.

Formally, the 5D-SLF is reformulated as ${L_{u,v}(x,y,\lambda)}$, which can be considered as an integration of 3D spectral data cubes across \(u \times v\) viewpoints. 
The phase plate imposes angle-dependent dispersion encoding on the light rays from different viewpoints, enabling the sensor to record angular information. The coding applied by the BCM results in the coupled coded field $A_{u,v}(x,y,\lambda) \cdot L_{u,v}(x,y,\lambda)$, with which $A_{u,v}(x,y,\lambda)$ denotes the filter function at viewpoint $(u,v)$.
The encoded SLF density integrated over all viewpoints is
\begin{equation}
{\bf{\hat{L}}}(x,y,\lambda) = \sum_{u,v \in U\times V } \!\!\!\!\!\!A_{u,v}(x,y,\lambda) \cdot L_{u,v}(x,y,\lambda),
\label{2-1-5}
\end{equation}
where \(U \times V\) denotes the aperture array corresponding to the phase plate.
Given the model of Eq. (\ref{2-1-3}), let $H$ be the filter of viewpoint $(u,v)$, and the transmittance of the sub-aperture $(u,v)$ be
\begin{equation}
A_{u,v}(x, y, \lambda) = H(\lambda;d_{u,v},\theta),
\label{2-1-6}
\end{equation}
where $d_{u,v}$ represents the thickness map of the birefringent quartz crystal corresponding to the viewpoint $(u,v)$. Thus, Eq. (\ref{2-1-5}) can be rewritten as:
\begin{equation}
{\bf{\hat{L}}}(x,y,\lambda) =\!\!\!\!\sum_{u,v \in U\times V }\!\!\!\!\!\!H(\lambda;d_{u,v},\theta) \cdot L_{u,v}(x,y,\lambda).
\label{2-1-7}
\end{equation}

Finally, when captured by an RGB sensor with a response function $R^c(\lambda)$, the encoded SLF cube ${\bf{\hat{L}}}$ becomes an RGB image measurement:
\begin{equation}
\setlength{\abovedisplayskip}{0.1cm}
\setlength{\belowdisplayskip}{0.1cm}
{I^{c \in {R,G,B}}}(x,y) = \int_{\lambda_1}^{\lambda_2}{\bf{\hat L}}(x,y,\lambda) \cdot R^c(\lambda) \, d\lambda + \eta(x,y),
\label{2-1-8}
\end{equation}
where $\lambda_1$ is the minimum wavelength, $\lambda_2$ is the maximum wavelength, and $\eta$ is sensor noise.

As described in Eq. (\ref{2-1-3}), the BCM can be used to modify the dispersion function by varying the angle $\theta$. Therefore, by recording measurements at different angles $\theta$, we can obtain multi-frame measurements to rich the modulated encoding.
Substituting Eq. (\ref{2-1-7}) into Eq. (\ref{2-1-8}), the multi-frame image formation of the ADLIS is described as
\begin{align}
{I_{\theta_i}^{c \in {R,G,B}}}(x,y) = \int_{\lambda_1}^{\lambda_2}\!&\left(\sum_{u,v}^{U\times V } \!\!H(\lambda;d_{u,v},\theta_i) \cdot L_{u,v}(x,y,\lambda)\right) 
\notag \\
& \cdot R^c(\lambda) \, d\lambda + \eta(x,y),
\label{2-1-9}
\end{align}
where ${I_{\theta_i}^{c \in {R,G,B}}}$ denotes the RGB image acquired from the $i$th frame measurement and $\theta_i$ represents the angle $\theta$ in frame $i$.

\subsection{Aperture-aware Dispersion Light-field Imaging (ADLI) Framework}\label{subsec:optimization}
To jointly optimize the optical encoder and the computational decoder, we introduce an E2E framework building upon previous work \cite{zhang2022end,baek2021single,ikoma2021depth,yang2024end,zheng2023close,shi2024split}.
Specifically, our Aperture-aware Dispersion Light-field Imaging (ADLI) framework consists of three key components - differentiable thickness-to-measurement generation, decoder network and loss function.

\begin{figure*}[t!]
\centering
\includegraphics[width=1\linewidth]{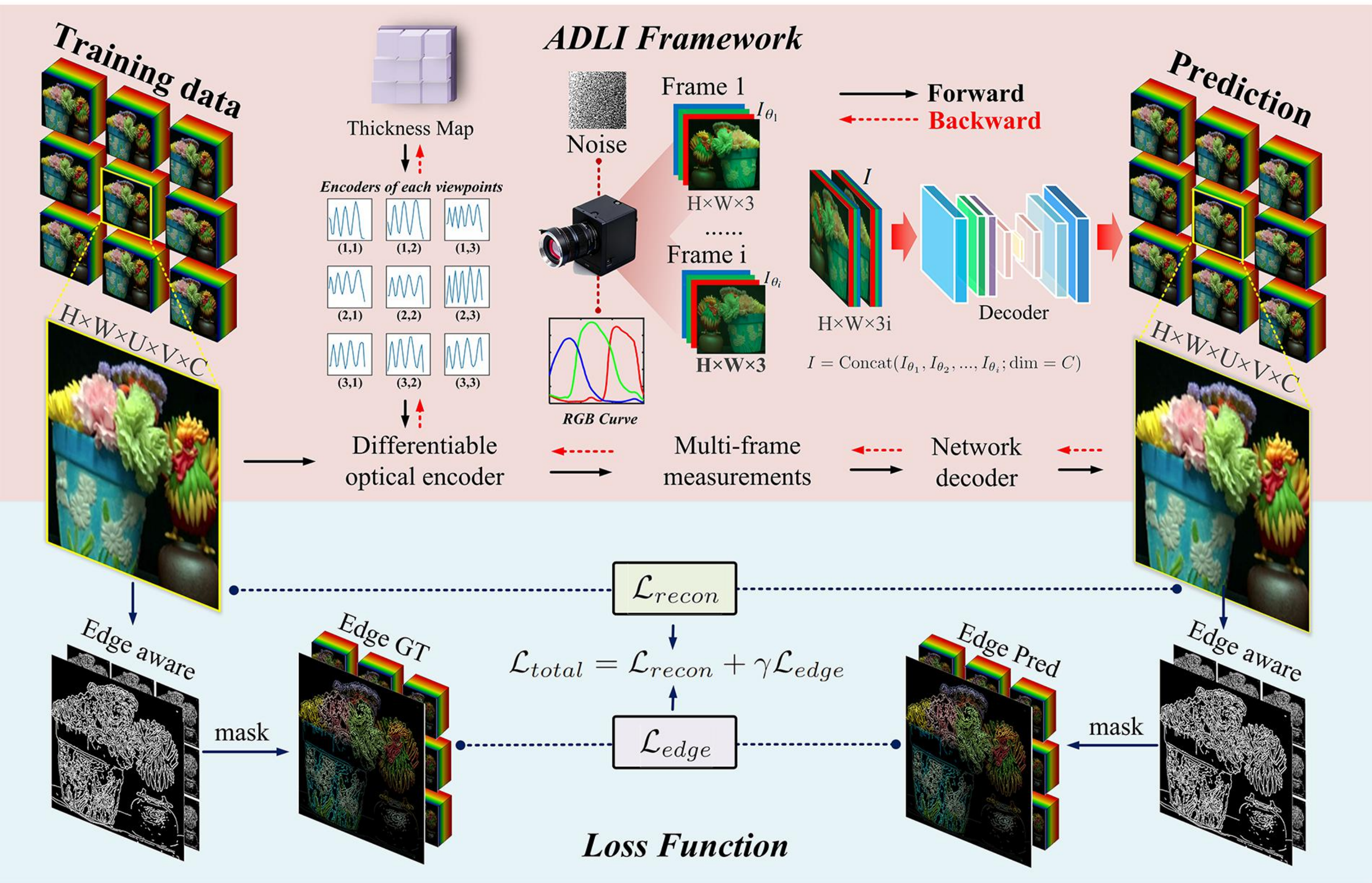}
  \caption{(Red) - Forward/backward propagation of the ADLI framework; (Blue) – Loss functions used.}
  \label{fig:pipeline}
\end{figure*}

\textbf{Differentiable thickness-to-measurement generation. }According to the aperture-aware dispersion ligh-field image formation model described in Section \ref{subsec:formation}, the thickness of the phase plate directly determines the modulation of the system.
Eq. (\ref{2-1-9}) establishes the relationship between the thickness map $d_{u,v}$ and intensity $I$ captured by the sensor. Therefore, the generation of measurement $I$ can be formulated concisely as
\begin{equation}
I_{\theta}(x,y) = \sum^{U\times V}_{u,v} \!\sum^{C_n}_{\lambda=C_1} \mathcal{H}_{\theta,\lambda}\left(d_{u,v}\right) \cdot L_{u,v,\lambda}(x,y)
\label{2-2-1}
\end{equation}
\begin{equation}
I = \mathrm{Concat}(I_{\theta_1},I_{\theta_2},...,I_{\theta_i}; \mathrm{dim}=C),
\label{2-2-2}
\end{equation}
where $L_{u,v,\lambda}(x,y)$ is an alternative representation of the 5D-SLF, while $\mathcal{H}_{\theta,\lambda}$ denotes the optical encoder parameterized by the differentiable variable $d_{u,v}$. 
And then, in Eq. (\ref{2-2-2}), $I\in\mathbb{R}^{H \times W \times 3i}$ is formed by concatenating the measurements $I_{\theta}\in\mathbb{R}^{H \times W \times 3}$ along the channel dimension, where each measurement is acquired under one of $i$ distinct analyzer angles $\theta$.
To model a fully differentiable encoder based on the phase plate, we formulate the physical model as a birefringent filter encoding model $\mathcal{F}$. 
Thus, the forward process of the encoder can be expressed as:
\begin{equation}
\bf{I} = \mathcal{F} \left( \bf{I}_{\text{raw}} ; d \right),
\label{2-2-1s}
\end{equation}
where $\bf{I}_{\text{raw}}$ equals to the original 5D-SLF ${L_{u,v}(x,y,\lambda)}$ and $\bf{d}$ equals to the $d_{u,v}$, which is learnable in the neural network.
To ensure the manufacturability and differentiability of optical components within the ADLI framework, we introduce an intermediate variable $\mathbf{w}_d$ to constrain the parameter $\textbf{d}$  within a feasible range. The constrained parameter is then defined as:
\begin{equation}
\mathbf{d} = d_{\text{min}} + \sigma(\mathbf{w}_d) \cdot (d_{\text{max}} - d_{\text{min}}),
\label{eq:param_constraint}
\end{equation}
where $\sigma$ denotes the sigmoid function. This formulation constrains $\mathbf{d}$ to the interval $(d_{\text{min}}, d_{\text{max}})$ while ensuring that the gradient $\frac{\partial \mathbf{d}}{\partial \mathbf{w}_d}$ exists and is continuous, which is crucial for the E2E differentiability of the pipeline. 
Specifically, the manufacturable range of the component is set from $d_{\text{min}} = 500\ \mu\text{m}$ to $d_{\text{max}} = 1000\ \mu\text{m}$. 
Consequently, when the encoded measurements $\mathbf{I}$, which are a function of the learnable parameter $\mathbf{d}$, are processed by the deep network, the parameters of the optical design can be updated in real-time via backpropagation as shown in Fig .\ref{fig:pipeline}. This mechanism enables the joint optimization of the optical encoder and the computational decoder.

To ensure viewpoint-specific encoding within practical manufacturing constraints, $d_{u,v}$ is initialized with nine preset random thickness values from view(1,1) to view(3,3): $644, 671, 692, 587, 683, 645, 905, 854, 597\ \mu\text{m}$. In our simulation setup, we configure $U \times V = 3 \times 3$, analyzer orientations $(\theta_1, \theta_2, \theta_3) = (45^\circ, 135^\circ, 195^\circ)$, and $C_n = 36$ spectral channels. It should be noted that the use of a $3 \times 3$ aperture array here represents a canonical implementation for prototype demonstration and should not be construed as the only or optimal configuration.

\textbf{Decoder network. } The Recovery of the 5D-SLF image from compressed measurements can be formulated as a convex optimization problem \cite{xiong2017snapshot,marquez2020compressive}. However, such optimization methods are typically time-consuming and suffer from severe under-determination. Therefore, to fully leverage the advanced datasets as physical constraints, we adopt the standard Restormer architecture \cite{zamir2022restormer} as the decoder, as shown in Fig. \ref{fig:pipeline}. In Section \ref{simulation}, results from various alternative decoder networks commonly used in image super-resolution and reconstruction tasks confirm the robustness of ADLIS to the choice of the decoder architecture.

\textbf{Loss functions. }The network receives multi-frame coded measurements as an input and outputs the recovered 5D-SLF ${\bf{\hat{Y}}}\in\mathbb{R}^{H \times W \times U \times V \times C}$. 
To recover the 5D-SLF more efficiently, we jointly optimize the learnable thickness map $d_{u,v}$ of the optics and the Restormer weights by minimizing the following loss function:
\begin{equation}
\mathcal{L}_{total} = \mathcal{L}_{recon}+\gamma\mathcal{L}_{edge},
\label{3-3-3}
\end{equation}
The training loss function $\mathcal{L}_{total}$ consists of the reconstruction loss $\mathcal{L}_{recon}$ on the 5D-SLF reconstruction and the edge-aware loss $\mathcal{L}_{edge}$:
\begin{equation}
\mathcal{L}_{recon} = {\left\| {{\bf{\hat{Y}}} - {\bf{Y}_{gt}}} \right\|^2},
\label{3-3-4}
\end{equation}
\begin{equation}
\mathcal{L}_{edge} = \sum^{U\times V}_{u,v} \!\sum^{C}_{\lambda}  \| E^{pred}_{u,v,\lambda} - E^{gt}_{u,v,\lambda} \|_1,
\label{3-3-6}
\end{equation}
where ${\left\| {\cdot} \right\|^2}$ denotes the L2 loss, ${\left\| {\cdot} \right\|_1}$ denotes the L1 loss, ${\bf{\hat{Y}}}$ and ${\bf{Y}_{gt}}$ are the predicted SLF and the ground truth, respectively.

In the edge-aware loss function, we reshape the 5D tensor ${\bf{\hat{Y}}}$ into 2D masks $Y(x,y)\in\mathbb{R}^{H \times W}$ by extracting each viewpoint and spectral band combination from the $U \times V$ viewpoints and $C$ spectral bands, and then apply the Sobel operator for each $Y(x,y)\in\mathbb{R}^{H \times W}$. 
Specifically, we apply the Sobel operator to both the reconstructed image $Y^{pred}_{u,v,\lambda}(x,y)$ and the ground truth image $Y^{gt}_{u,v,\lambda}(x,y)$, and compute the horizontal and vertical gradient responses:

\begin{equation}
E^{pred}_{u,v,\lambda} = \sqrt{(Y^{pred}_{u,v,\lambda} * S_x)^2 + (Y^{pred}_{u,v,\lambda} * S_y)^2}
\end{equation}
\begin{equation}
E^{gt}_{u,v,\lambda} = \sqrt{(Y^{gt}_{u,v,\lambda} * S_x)^2 + (Y^{gt}_{u,v,\lambda} * S_y)^2},
\end{equation}
where $S_x$ and $S_y$ are the Sobel kernels in the horizontal and vertical directions respectively, and $*$ denotes the convolution operation. Actually, the disparity features in a light field are often concentrated along sharp edges and areas rich in geometric detail. In contrast, large smooth regions usually exhibit no significant variation across different viewpoints. As a result, the edge loss function $\mathcal{L}_{edge}$  enhances the structural distinction between the viewpoints, thereby improving the quality of reconstruction.

\section{SIMULATIONS}\label{simulation}

\subsection{Implementation Details}\label{subsec:SET}
We adopt the recently published 5D-SLF dataset, RealSLF \cite{li2025realslf}, for simulations. The dataset contains  $7 \times 5$ viewpoints with 36 spectral bands from 450 to 750 nm at 8.33 nm intervals. To match the $3 \times 3$ aperture array of our system, we crop patches of size $400 \times 400 \times 3 \times 3 \times 36$ in size from the original 5D-SLF cubes for training. The dataset of 25 carefully curated scenes was split into 19 scenes for training and 6 for testing.
The thickness map of the phase plate and the reconstruction network are trained in an E2E pipeline. We train the entire model for 200 epochs with a batch size of eight and a patch size of $50 \times 50$ using Adam optimizer. We set $\gamma$ = 0.001 in Eq. (\ref{3-3-3}) to constrain the loss function to balance. The initial learning rate is 0.002 for the optical encoder and 0.0001 for the computational decoder, both with a decay rate of 0.9 per epoch. The phase plate optimization commences at epoch 5 and ends at epoch 150. All the models are implemented using PyTorch and trained on an NVIDIA GeForce RTX 4090 GPU. 
To evaluate the quality of the reconstruction results, we employ peak signal to noise ratio (PSNR), structural similarity (SSIM) \cite{smith1931cie} and spectral angular mapper (SAM) \cite{kruse1993spectral}. 



\subsection{Comparison with Different Encoding Elements }
To validate the superiority of the selection of BCM as the encoding element, and the E2E ADLI framework, we first assess the imaging performance of three aperture-encoded optical models: (1) a conventional color-filter-array-based (CFA-based) encoding solution \cite{lopez2024depth,zhu2018hyperspectral,marquez2020compressive,zhang2024compact,monakhova2020spectral}; (2) a reference ADLI model with fixed design parameters; and (3) the proposed ADLI framework using our E2E paradigm where optical parameters are learnable. For each model, the reconstruction results using 1, 2 and 3 measurement frames are presented. 

Specifically, for the CFA-based model used in our comparative study, we simulate the spectral encoding properties of a $3 \times 3$ color filter array using Gaussian transmission functions. The central wavelengths of the nine Gaussian filters are equally spaced across the spectral range from 450 nm to 750 nm, providing comprehensive coverage of the visible spectrum.

The transmission function for each filter element follows a Gaussian distribution:
\begin{equation}
T_i(\lambda) = \exp\left(-\frac{(\lambda - \mu_i)^2}{2\sigma_i^2}\right),
\end{equation}
where $\mu_i$ denotes the central wavelength of the $i$-th filter, equally distributed between 450--750 nm, and $\sigma_i$ represents the bandwidth parameter randomly sampled between 10--35 nm to simulate realistic filter characteristics with varying spectral selectivity. The complete CFA encoding process involves multiplying the incident spectral light field with these transmission functions, followed by integration with standard RGB sensor response curves. The RGB responses are modeled as Gaussian functions centered at 480 nm (blue), 550 nm (green), and 680 nm (red) with 80 nm bandwidth:
\begin{equation}
R^c(\lambda) = \exp\left(-\frac{(\lambda - \mu_c)^2}{2\sigma_c^2}\right), \quad c \in \{\text{R}, \text{G}, \text{B}\},
\end{equation}
where $\mu_c$ represents the central wavelength for each color channel and $\sigma_c = 80$ nm. This simulation framework provides a physically-grounded representation of conventional CFA-based spectral imaging systems for fair comparison with our proposed ADLIS approach.
To make the comparison fair, we adopt the same reconstruction network, Restormer \cite{zamir2022restormer}, as the decoder for all optical encoding models. 
Meanwhile, all models are trained for 200 epochs with the same optimizer configuration.

\begin{table*}[!htb]
\centering
\caption{The numerical comparison of the reconstruction results.}
\renewcommand{\arraystretch}{1}  
\resizebox{\textwidth}{!}{
\fontsize{5}{6}\selectfont  
\begin{tabular}{cccccccccc}
\hline
\rowcolor[HTML]{FFFFFF} 
\textbf{Encoding}               
& \multicolumn{3}{c}{\textbf{PSNR} $\uparrow$}  
& \multicolumn{3}{c}{\textbf{SSIM} $\uparrow$}                           
& \multicolumn{3}{c}{\textbf{SAM} $\downarrow$}            \\ \hline
Frame(s) for reconstruction 
& \textbf{1} 
& \textbf{2}                            
& \textbf{3}                            
& \textbf{1} 
& \textbf{2}       
& \textbf{3}                             
& \textbf{1} 
& \textbf{2}       
& \textbf{3}                            
\\ \hline
CFA (baseline)                             
& 35.07      
& \textbackslash{}                      
& \textbackslash{}                      
& 0.9586
& \textbackslash{} 
& \textbackslash{}                       
& 11.38      
& \textbackslash{} 
& \textbackslash{} \\
ADLI-fixed                      
& 37.02    
& 39.24                                
& 40.55                               
& 0.9645    
& 0.9764         
& 0.9811                                
& 7.78     
& 7.44         
& 7.33 \\
ADLI-learnable (proposed)               
& 38.65     
& {\color[HTML]{000000} \textbf{41.36}} 
& {\color[HTML]{CB0000} \textbf{42.07}} 
& 0.9723   
& \textbf{0.9817}  
& {\color[HTML]{CB0000} \textbf{0.9839}} 
& 7.58      
& \textbf{6.85}   
& {\color[HTML]{CB0000} \textbf{6.63}} \\ \hline
\end{tabular}}
\label{tab1}
\end{table*}
\begin{figure*}[!htb]
\centering
\includegraphics[width=1\linewidth]{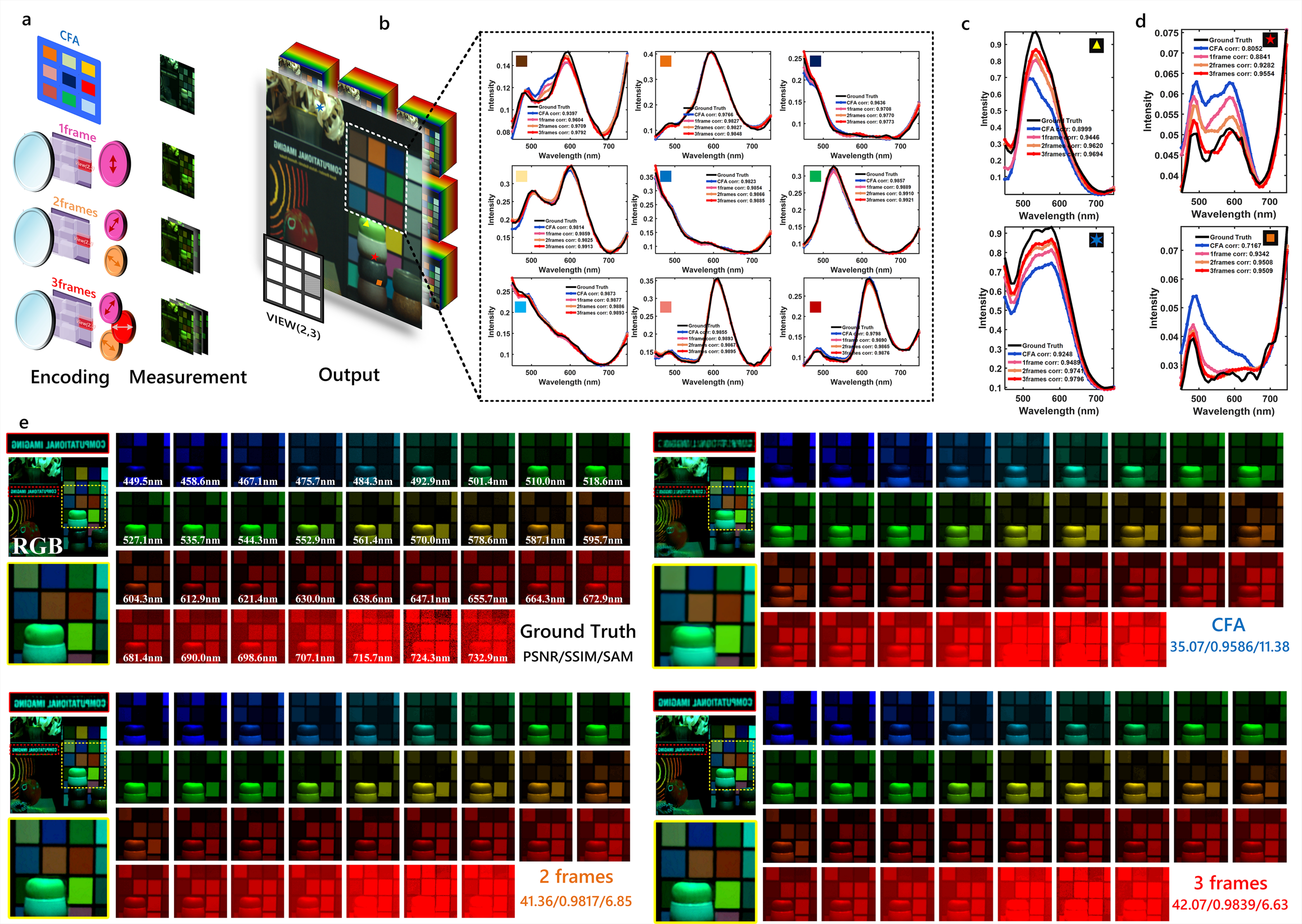}
\caption{The reconstruction results of view (2,3) in a test scene. (a) Encoding models and their corresponding measurements. (b) The recovered 5D-SLF cube and the spectral curves of the ROI (color-checker) in the scene. (c) The spectral curves of scene points approaching overexposure. (d) Simulation results compared with the GT. PSNR, SSIM and SAM results are also shown for each encoding model. (e) The spectral curves of scene points approaching underexposure.}
\label{fig:spectral}
\end{figure*}

Table \ref{tab1} presents the average PSNR, SSIM, and SAM values for the test set. We observe that both ADLI models outperform the conventional model, demonstrating the superiority of BCM for 5D-SLF modulation and imaging. As shown in Table \ref{tab1}, the results consistently improve when more frames are used as input to the ADLI models, demonstrating that ADLIS can enhance the imaging performance through multi-frame measurements.
Furthermore, the ADLI model with learnable parameters achieves a better performance, validating the effectiveness of the proposed E2E joint optimization design.

Notably, although 3-frame acquisition shows some improvement over 2-frame acquisition, we select the 2-frame model for the final ADLIS prototype because it allows us to strategically choose two analyzer angles $\theta_1, \theta_2$ such that the second term of Eq. \eqref{2-1-2} ($ \sin2\alpha\sin2\theta $) becomes opposite for the two measurements, thereby maximizing the encoding disparity. This 2-frame configuration ensures maximum encoding difference between frames, while extra frames could theoretically improve the reconstruction quality slightly through an extra encoding function which only differs from the previous two frames in terms of amplitude, we do not do this to avoid increasing complexity. Fig. \ref{fig:pola} shows the encoding response functions corresponding to the two measurement frames.

\subsection{Spectral Performance}
To evaluate the spectral accuracy of the reconstructions, we present the reconstruction results of view (2,3) in a test scene in Fig. \ref{fig:spectral}.
Using the CFA-based model as the baseline, we qualitatively compare the spectral reconstruction performance of the proposed ADLIS.
The measurement encoding processes of different models are illustrated in Fig. \ref{fig:spectral}a. We also present the spectral curves of the points on a color checker, high-intensity scene points, and low-intensity scene points, respectively in Fig. \ref{fig:spectral}c, and \ref{fig:spectral}d.
All curves were reconstructed using each encoding model, with the correlation coefficients calculated against the ground truth (GT).
Clearly, the spectral curves reconstructed by the proposed ADLI framework exhibit superior agreement with the GT.

Notably, when more frames of encoded measurements are employed, the spectral reconstruction accuracy for scene points approaching the overexposed (positions marked by the yellow triangles and blue hexagrams in Fig. \ref{fig:spectral}b) or underexposed (positions marked by the orange square and red pentagram in Fig. \ref{fig:spectral}b) regions show greater improvement. 
The proposed ADLIS leverages the distinct polarization properties of light from different scene regions including strong specular highlights and weak diffuse reflections by rotating a polarizer to capture multiple frames under different polarization states. Each frame is sensitive only to some specific polarization directions, allowing the system to recover complete intensity information with highlight details preserved and shadow regions discernible. This indicates that ADLIS is inherently capable of achieving a high dynamic range.
Examination of the magnified region of interest (ROI) reveals that the ADLI framework produces visually appealing results with reduced artifacts and sharper edges compared to the baseline. Fig. \ref{fig:spectral}e shows 34 of the 36 reconstructed spectral bands from a RealSLF dataset scene for each encoding model. The ADLI framework with three measurement frames yields the best results, with reconstructed images showing fine details and sharp edges, whereas the CFA-based model exhibits block-like artifacts. This experiment validates the spectral imaging capability of the proposed system. 

To further verify the outperformance in spectral encoding of the proposed ADLI framework, we conduct extended simulations to compare the reconstruction results of proposed model with conventional model in different numbers of spectral bands selected from the dataset, including 25 bands (493-699$nm$), 31 bands (467-724$nm$) and the original 36 bands (450-750$nm$).
Table \ref{tab2} summarizes the average results of the test scenes.
Typically, when reconstructing more spectral bands, the model can leverage the correlations between bands in the hyperspectral data as a constraint, which consequently demonstrates gains in metrics such as PSNR, SSIM, and SAM.
Notably, it can be seen that in the $\mathbf{\Delta}$sam metric between ADLI framework and conventional CFA-based model under these 3 band settings, whose results demonstrate that as the number of spectral bands to be reconstructed increases, the superiority of the ADLI framework on the $\mathbf{\Delta}$sam metric becomes more pronounced, indicating the inherent advantage of the ADLIS encoding model in spectral dimension.

\begin{table}[!tb]
\centering
\caption{Spectral reconstruction performance of the proposed ADLI framework under different numbers of spectral bands. The conventional CFA-based model is used as a baseline.}
\renewcommand{\arraystretch}{1.2}  
\setlength{\tabcolsep}{2pt}  
\resizebox{80mm}{!}{
\fontsize{6}{7}\selectfont
\begin{tabular}{@{}cccccc@{}}  
\hline
\rowcolor[HTML]{FFFFFF} 
\textbf{Spectral Channel} & \textbf{Encoding} & \textbf{PSNR} & \textbf{SSIM} & \textbf{SAM} & $\mathbf{\Delta}$sam \\ \hline  
\multirow{4}{*}{25 Bands}                       
& CFA                                    
& 34.32                              
& 0.9602                             
& 9.48                              
& \textbf{0}                
\\    
& 1 Frame                                
& 37.32
& 0.9645                           
& 7.06                    
& \textbf{2.42}                     
\\                     
& 2 Frames                               
& 38.46                           
& 0.9726                             
& 7.03                         
& \textbf{2.45}                     
\\ 
& 3 Frames                               
& 40.49             
& 0.9817                             
& 6.99                            
& \textbf{2.49}                     
\\ \hline
\multirow{4}{*}{31 Bands}                       
& CFA                                    
& 35.71                             
& 0.9654                   
& 9.70                          
& \textbf{0}                         
\\ 
& 1 Frame                                
& 37.17                           
& 0.9661                            
& 7.10                             
& \textbf{2.60}                     
\\ 
& 2 Frames                               
& 39.22                           
& 0.9742                       
& 7.09                    
& \textbf{2.61}                     
\\ 
& 3 Frames                               
& 40.17                        
& 0.9795          
& 6.82                           
& \textbf{2.88}                     
\\ \hline
\multirow{4}{*}{36 Bands}                       
& CFA                                    
& 35.07                      
& 0.9586                          
& 11.38                       
& \textbf{0}                         
\\ 
& 1 Frame                                
& 38.65                            
& 0.9723                            
& 7.58                             
& \textbf{3.80}                     
\\ 
& 2 Frames                               
& 41.36       
& 0.9817                  
& 6.85                             
& \textbf{4.53}                     
\\ 
& 3 Frames                               
& 42.07                           
& 0.9839                        
& 6.63
& \textbf{4.75}                     
\\ \hline
\end{tabular}}
\label{tab2}
\end{table}

\begin{table}[!tb]
\centering
\caption{The angular reconstruction performance.}
\renewcommand{\arraystretch}{1.2}  
\setlength{\tabcolsep}{2pt}  
\resizebox{88mm}{!}{
\begin{tabular}{@{}ccccccccc@{}}  
\hline
\rowcolor[HTML]{FFFFFF} 
\textbf{}      
& \multicolumn{4}{c}{\textbf{Pearson Corr} $\uparrow$}  
& \multicolumn{4}{c}{\textbf{RMSE} $\downarrow$}       \\ \hline
\textbf{Scene} 
& CFA                                    
& 1 Frame                                
& 2 Frames                               
& 3 Frames                                                    
& CFA                                    
& 1 Frame                                
& 2 Frames                               
& 3 Frames           
\\ \hline
\textbf{S1}    
& {\color[HTML]{000000} 0.7600}          
& 0.7319                                 
& {\color[HTML]{000000} \textbf{0.7782}} 
& \multicolumn{1}{c|}{{\color[HTML]{CB0000} \textbf{0.8961}}} 
& {\color[HTML]{000000} \textbf{0.1581}} 
& {\color[HTML]{000000} 0.1913}          
& {\color[HTML]{CB0000} \textbf{0.1486}} 
& {\color[HTML]{000000} 0.1695}          \\
\textbf{S2}    
& {\color[HTML]{000000} \textbf{0.9062}} 
& {\color[HTML]{000000} 0.8866}          
& 0.8869                                 
& \multicolumn{1}{c|}{{\color[HTML]{CB0000} \textbf{0.9073}}} 
& {\color[HTML]{000000} 0.1993}          
& {\color[HTML]{000000} 0.1740}          
& {\color[HTML]{000000} \textbf{0.1569}}
& {\color[HTML]{CB0000} \textbf{0.1334}} \\
\textbf{S3}    
& 0.7930                                 
& {\color[HTML]{CB0000} \textbf{0.8862}} 
& {\color[HTML]{000000} \textbf{0.8533}} 
& \multicolumn{1}{c|}{{\color[HTML]{000000} 0.8339}}          
& {\color[HTML]{000000} 0.1053}          
& {\color[HTML]{000000} \textbf{0.0795}} 
& {\color[HTML]{000000} 0.1078}          
& {\color[HTML]{CB0000} \textbf{0.0568}} \\
\textbf{S4}    
& {\color[HTML]{000000} 0.9490}          
& {\color[HTML]{000000} \textbf{0.9571}} 
& 0.9307                                 
& \multicolumn{1}{c|}{{\color[HTML]{CB0000} \textbf{0.9745}}} 
& {\color[HTML]{000000} \textbf{0.1203}} 
& {\color[HTML]{000000} 0.1506}          
& {\color[HTML]{000000} 0.1360}          
& {\color[HTML]{CB0000} \textbf{0.1059}} \\
\textbf{S5}    
& 0.9171                                 
& {\color[HTML]{000000} 0.9108}          
& {\color[HTML]{CB0000} \textbf{0.9513}} 
& \multicolumn{1}{c|}{{\color[HTML]{000000} \textbf{0.9404}}} 
& {\color[HTML]{000000} 0.0856}          
& {\color[HTML]{000000} 0.0740}          
& {\color[HTML]{000000} \textbf{0.0621}}
& {\color[HTML]{CB0000} \textbf{0.0576}} \\
\textbf{S6}    
& {\color[HTML]{000000} \textbf{0.8444}} 
& 0.7944                                 
& {\color[HTML]{000000} 0.7782}          
& \multicolumn{1}{c|}{{\color[HTML]{CB0000} \textbf{0.8961}}} 
& {\color[HTML]{000000} 0.2608}          
& {\color[HTML]{000000} 0.1729}          
& {\color[HTML]{CB0000} \textbf{0.0851}} 
& {\color[HTML]{000000} \textbf{0.0922}} \\ \hline
\textbf{Avg.}  
& 0.8616                                 
& 0.8612                                 
& {\color[HTML]{000000} \textbf{0.8631}} 
& {\color[HTML]{CB0000} \textbf{0.9081}}                      
& 0.1549                                 
& 0.1404                                 
& {\color[HTML]{000000} \textbf{0.1161}} 
& {\color[HTML]{CB0000} \textbf{0.1026}} \\ \hline
\end{tabular}}
\label{tab4}
\end{table}

\begin{figure}[!ht]
\centering
\includegraphics[width=1\linewidth]{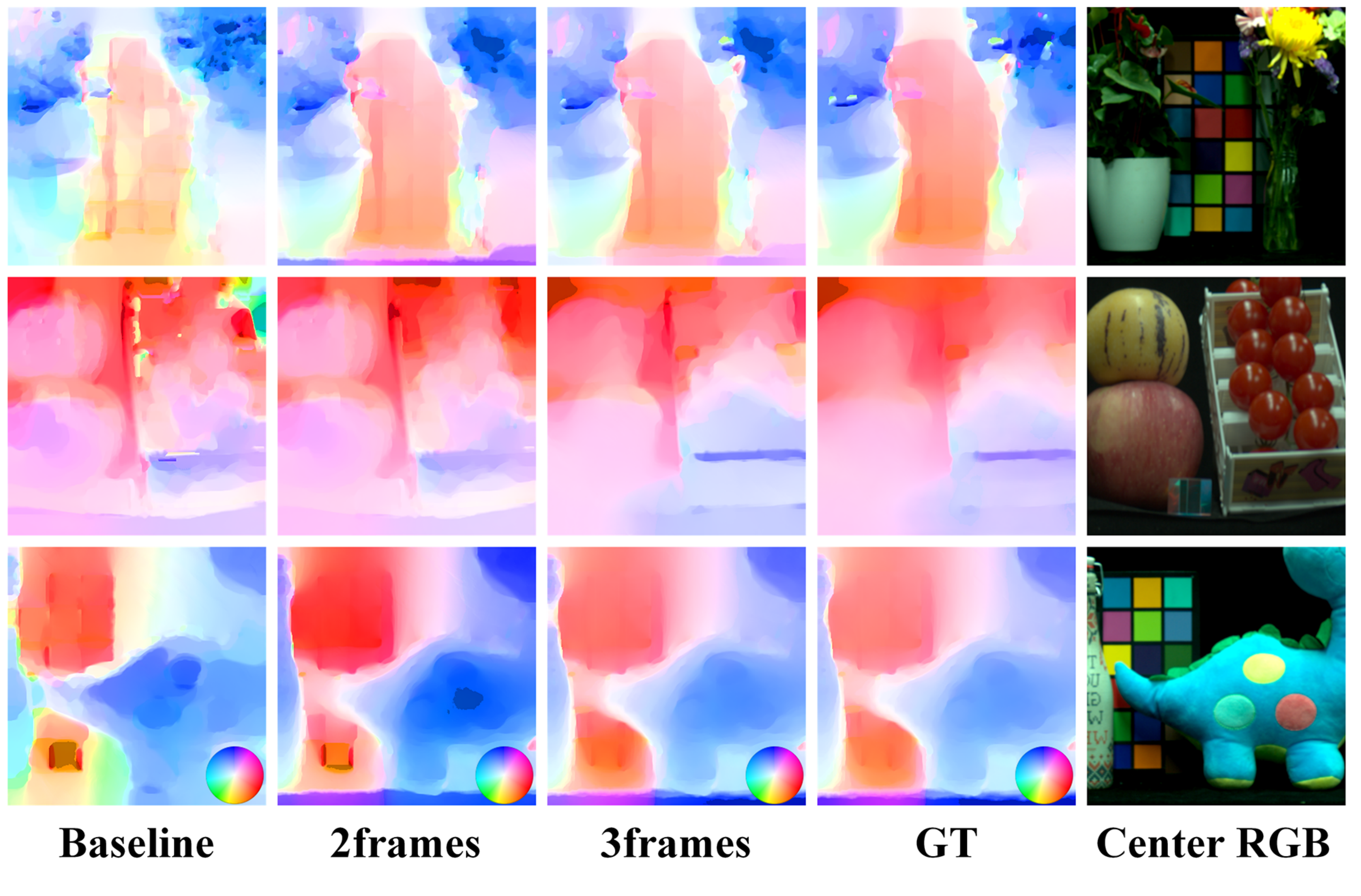}
  \caption{The disparity maps estimated from the original dataset (GT), the proposed ADLI framework (2frames and 3frames) and the CFA-based model (Baseline). }
  \label{fig:depth}
\end{figure}

\subsection{Angular Performance}
To evaluate the angular reconstruction quality quantitatively and qualitatively, we present the simulation results in Table \ref{tab4} and Fig. \ref{fig:depth}.
Due to the absence of absolute depth information in the RealSLF dataset, which provides only multi-view images, we employ relative disparity maps as a reference for evaluating angular reconstruction performance. These maps are estimated from multi-view correspondences according to \cite{ipol.2013.26} and normalized to [0, 1], where 0 and 1 correspond to the nearest and farthest points from the imaging system, respectively.

\begin{table*}[!ht]
\centering
\caption{The reconstruction results of different algorithms.}
\renewcommand{\arraystretch}{1}  
\resizebox{\textwidth}{!}{
\fontsize{10}{13}\selectfont  
\begin{tabular}{ccccccccccc}
\hline
\textbf{Algorithm}  & GAP-Net & U-Net    & HSCNN-Plus & HRNet  & $S^2$-Transformer & HDNet  & EDSR   & MIRNet & MPRNet & Restormer \\
\textbf{Reference}  & IJCV2023 & MICCAI2015   & CVPR2018 & CVPRW2020  & TPAMI2025 & CVPR2022  & CVPR2017   & ECCV2020 & CVPR2021 & CVPR2022
\\ \hline
\textbf{PSNR  $\uparrow$ }     & 31.86   & 35.09  & 37.28      & 37.73  & 37.82           & 38.03  & 38.22  & 38.57  & 39.82  & 41.36     \\
\textbf{SSIM  $\uparrow$ }     & 0.9012  & 0.9187 & 0.9644     & 0.9674 & 0.9150          & 0.9690 & 0.9704 & 0.9718 & 0.9628 & 0.9817    \\
\textbf{SAM $\downarrow$ }     & 23.78   & 19.68  & 8.58       & 8.15   & 12.83           & 7.87   & 7.65   & 8.51   & 12.58  & 6.85      \\ \hline
\textbf{Params(M)} & 4.30    & 1.04   & 1.28       & 5.14   & 0.26            & 1.40   & 2.82   & 9.11   & 4.66   & 3.98      \\
\textbf{FLOPs(G)}  & 47.12   & 2.16   & 51.08      & 18.81  & 11.88           & 55.77  & 112.70 & 74.25  & 163.76 & 20.57     \\ \hline
\end{tabular}}
\label{tab4.5}
\end{table*}

\begin{figure*}[t]
\centering
\includegraphics[width=1\linewidth]{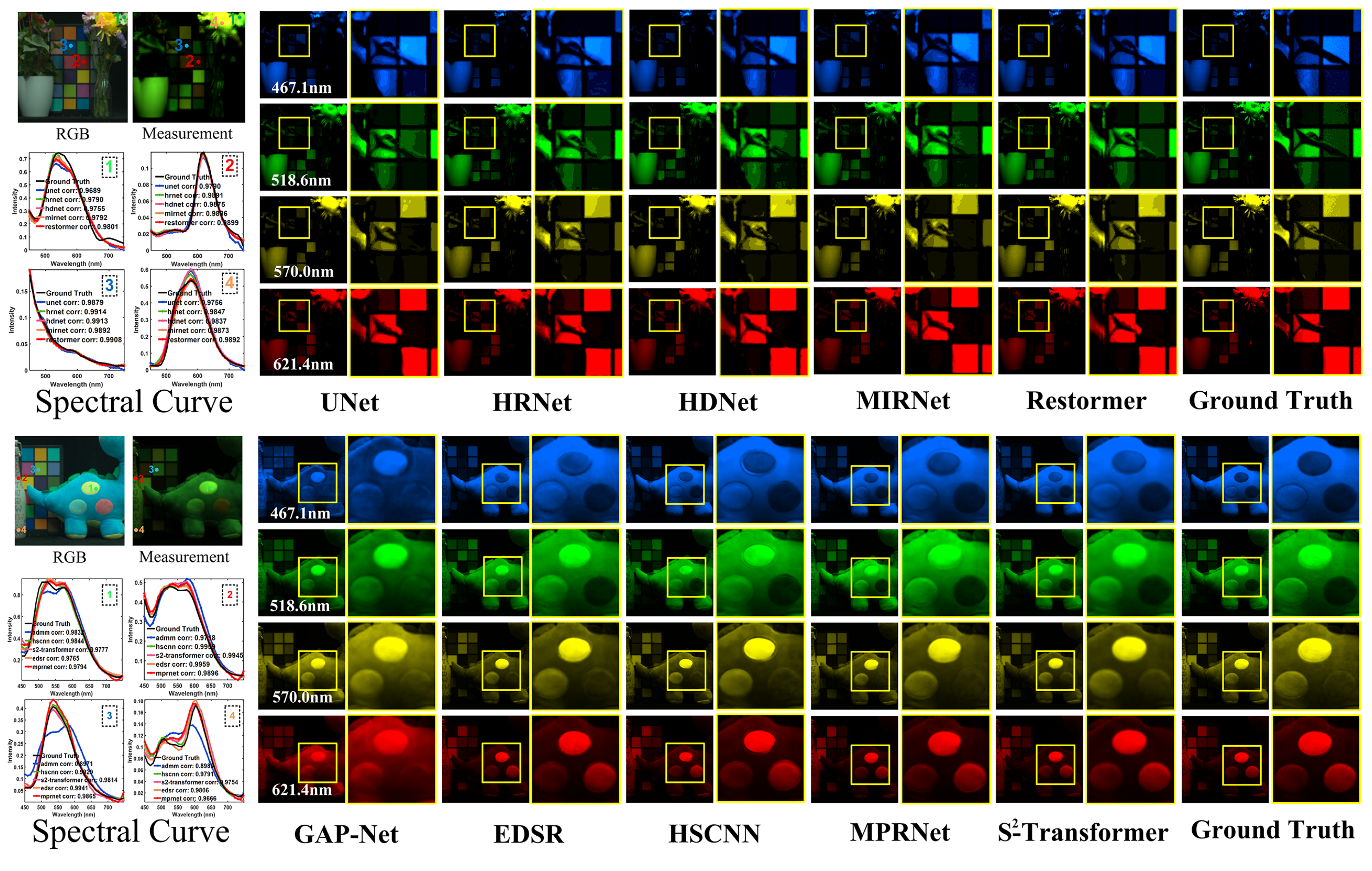}
  \caption{Simulation results of different algorithms. We present the reconstruction results of the central view at four representative wavelengths, along with spectral curves from selected scene points. }
  \label{fig:spectral2}
\end{figure*}

Specifically, Fig. \ref{fig:depth} presents the estimated disparity maps for test scenes from RealSLF. Compared with the conventional CFA-based baseline, the proposed ADLI model significantly improves the disparity-estimation accuracy in our test scenes.
To further evaluate the angular reconstruction performance quantitatively, Table \ref{tab4} reports the Pearson correlation coefficient (Pearson Corr) and root mean squared error (RMSE) between the ground truth depth maps and those obtained using different encoding models. The results indicate that methods using 3-frame measurements produce more accurate disparity estimates than their fewer-frame counterparts and the conventional CFA-based baseline.


\subsection{Simulation Results of Different Algorithms}

Owing to the lack of research on learning-based 5D-SLF reconstruction, we present the reconstruction results based on classic learning-based methods, including GAP-Net \cite{meng2023deep}, UNet \cite{ronneberger2015u}, HSCNN+ \cite{shi2018hscnn+}, HRNet \cite{zhao2020hierarchical}, $S^2$-Transformer \cite{wang2025sˆ}, HDNet \cite{hu2022hdnet}, EDSR \cite{lim2017enhanced}, MIRNet \cite{zamir2020learning}, MPRNet \cite{zamir2021multi} and Restormer \cite{zamir2022restormer}, to verify the generalizability of the proposed E2E ADLIS-2frames model across various decoders as an example. 
Table \ref{tab4.5} presents the simulation results for different reconstruction algorithms. All these algorithms remain the default implementation and model architecture without any refinement.
Overall, most algorithms achieve a PSNR of 35 dB, SSIM of 0.96, demonstrating that our imaging system delivers excellent reconstruction performance across various established networks. This result underscores its strong generalizability and eliminates the need for customized learning-based reconstruction networks. Among all evaluated networks, Restormer achieves the best performance across all metrics and is therefore selected as the decoder in the proposed ADLI framework, as shown in Fig. \ref{fig:pipeline}. Furthermore, the parameter count and FLOPs are reported to validate a balanced model design, confirming that performance improvements stem from an optimized architecture rather than an artificially simplified or overcomplicated network. Fig \ref{fig:spectral2} shows the visualized results of the center view predicted by different algorithms. We find out that all other methods performed well except for Gap-Net.


\begin{figure}[!htb]
\centering
\includegraphics[width=1\linewidth]{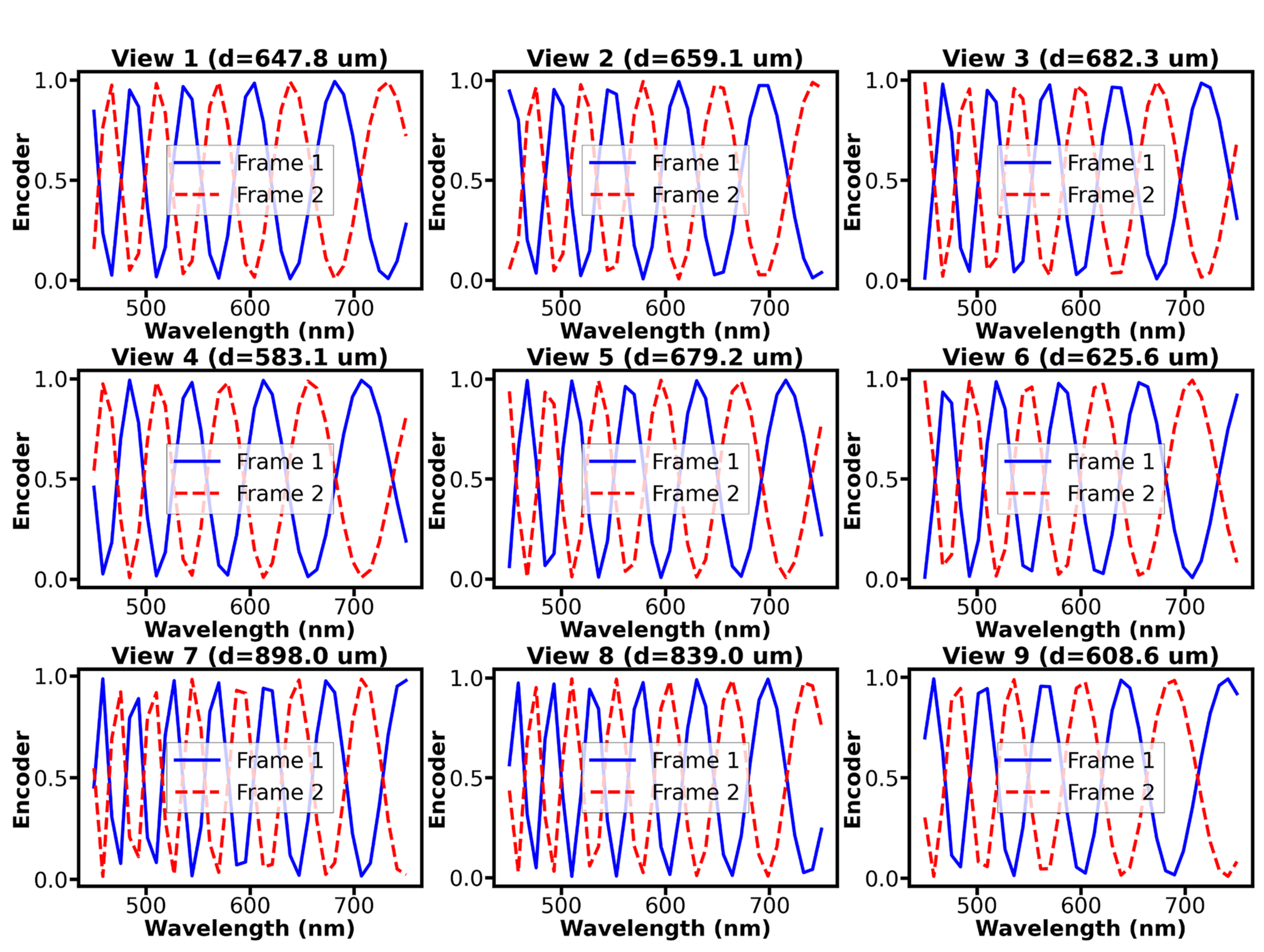}
  \caption{The encoding response function of 2-frames ADLI framework. }
  \label{fig:response}
\end{figure}

\section{EXPERIMENTS ON REAL DATA}
Building on the reasonable encoding modulation and outstanding reconstruction quality demonstrated in Section \ref{simulation}, we select the ADLIS-2frames model for experimental validation with real data.
Following the ADLI-2frames framework, the thickness maps for views(1,1) to view(3,3) are ultimately converged to 647.8, 659.1, 682.3, 583.1, 679.2, 625.6, 898.0, 839.0 and 608.6 $\mu\text{m}$, respectively. The final encoding function of the aperture for 3 × 3 views above is presented in Fig. \ref{fig:response}.
It can be observed that the red curve and the blue curve can theoretically achieve maximum disparity between 2 frames, i.e., they are opposites of each other.
Using this optimized thickness profile, we fabricate an ADLIS prototype and conduct real-world experiments.

\begin{figure}[!htb]
\centering
\includegraphics[width=1\linewidth]{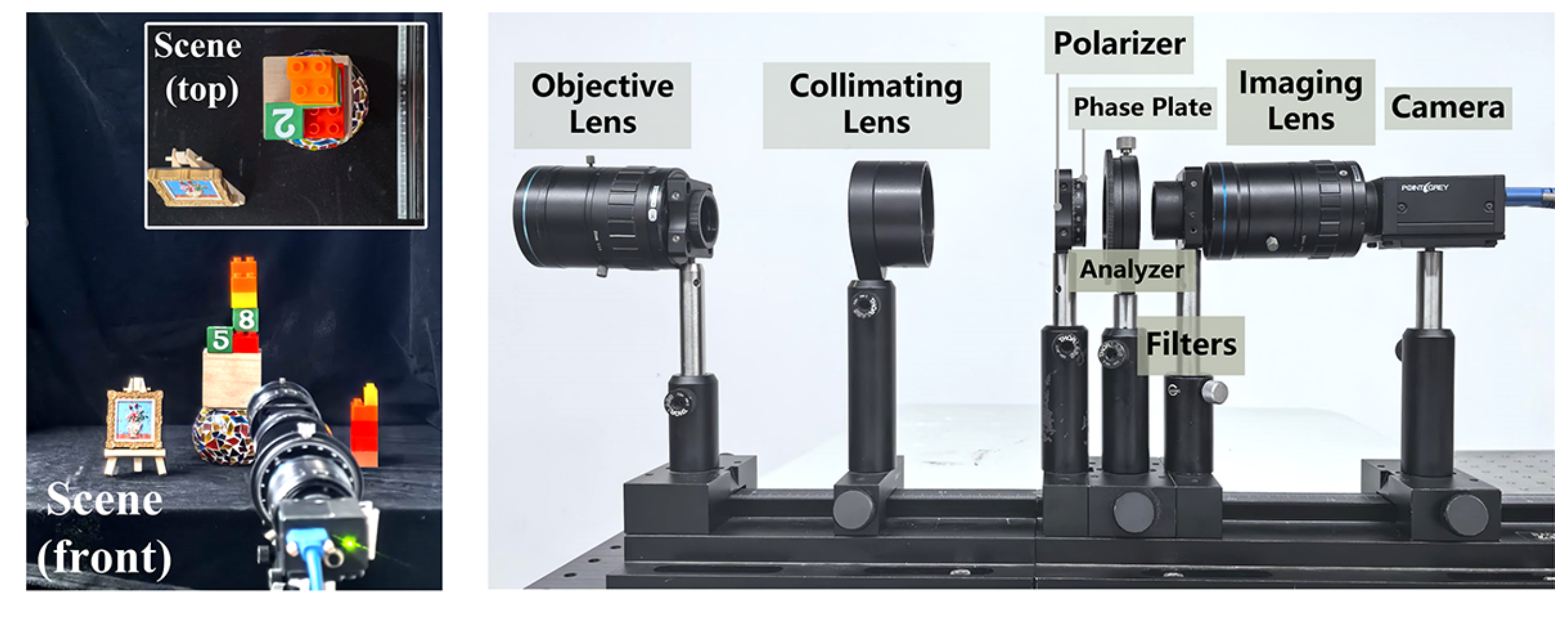}
  \caption{We arrange a spectrally interesting and texturally rich test scene (Left) to evaluate the real-world performance of the ADLIS prototype. Implementation of hardware system (Right). }
  \label{fig:system}
\end{figure}

\subsection{Prototype}
\begin{figure*}[!htb]
\centering
\includegraphics[width=1\linewidth]{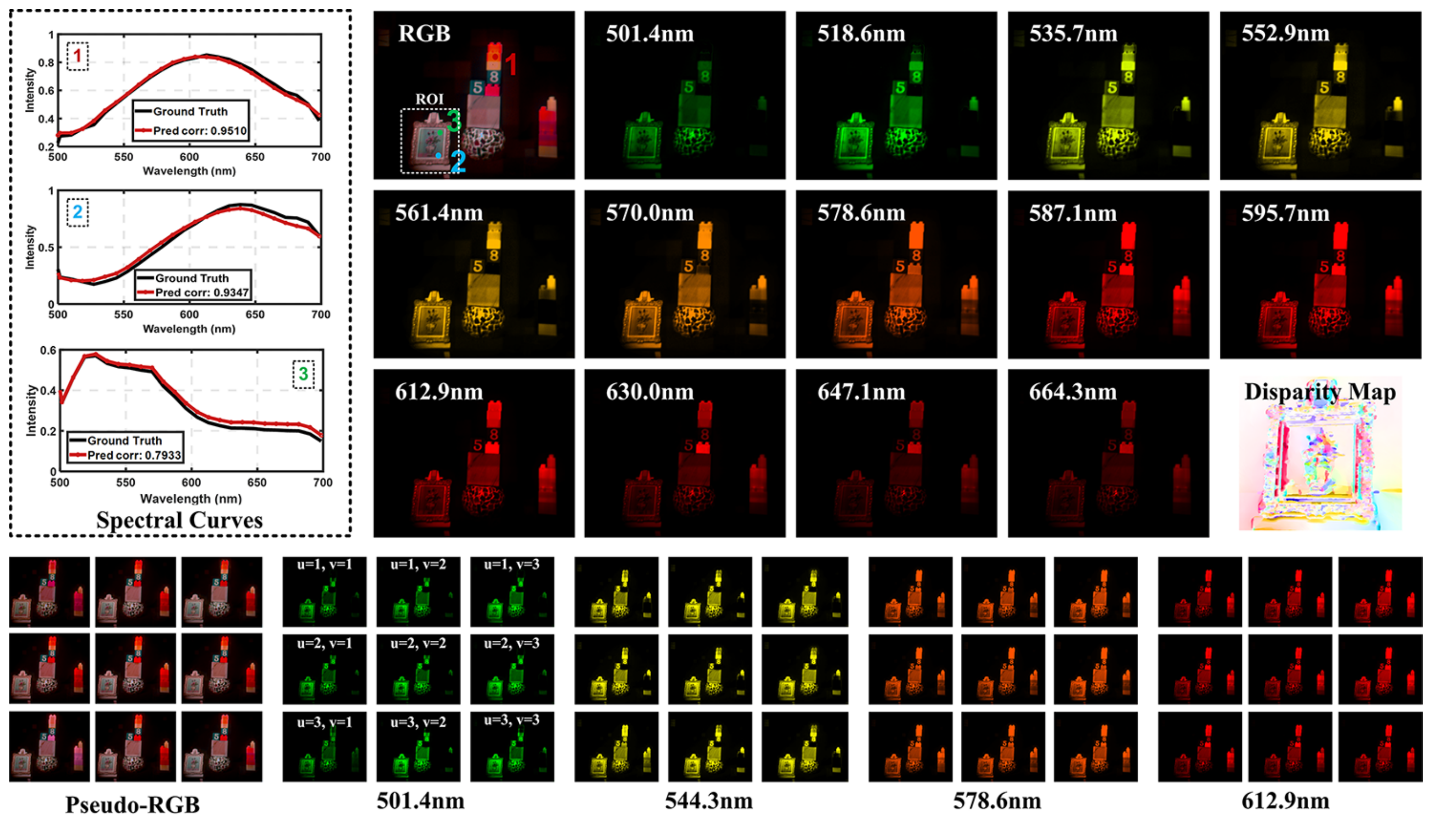}
  \caption{ We develop the ADLI framework to recover 5D-SLF from 2D compressed measurements. Spectral performance of ADLIS in real-world experiments (top). We select the center viewpoint for presentation. Here, we also choose a region of interest to show the disparity map. (Bottom) Angular performance of ADLIS in real-world experiments. }
  \label{fig:real}
\end{figure*}

As illustrated in Fig. \ref{fig:system}, the hardware system consists of a 25$mm$ objective lens (Myutron HF5018V F1.8), a 100$mm$ collimating lens (Thorlabs AC508-100-A-ML), two polarizers, a co-optimized phase plate made of quartz crystal, a diaphragm and a detector (Point Grey GS3-U3-51S5C). The detector has a resolution of 2448 × 2048 pixels with a pixel pitch of 3.45$\mu$$m$. This prototype also includes additional filters with a wavelength range of 500$nm$ to 700$nm$ (LBTEK MEFH10-500LP and MEFH10-700SP) to restrict the operating band. The phase plate has a diameter of 25.4$mm$ and includes an 18$mm$ × 18$mm$ modulation surface.
To address the domain gap between the simulation and real-world experiments caused by inherent noise and unmodeled optical fabrication errors, we augment the training data with randomized noise during model training.
For performance evaluation, we image a spectrally and angularly rich scene containing a colored painted relief and several building blocks, as shown in Fig. \ref{fig:real} (top left). The exposure time is set to 30 ms under D65 illumination, which is consistent with the RealSLF dataset configuration. Using two edge-pass filters to limit the spectral range to 500–700 nm, we reconstruct 25 spectral bands as the final outputs through wavelength-dependent dispersion calibration.

\subsection{Captured Results}
The spectral performance of the real 5D-SLF reconstruction is shown in Fig. \ref{fig:real}. The proposed ADLIS capture measurements with a resolution of 2448 × 2048 pixels from real-world scenes and recover the 5D-SLF in the size of 2448 × 2048 × 3 × 3 × 25 computationally in an E2E framework. Fig. \ref{fig:real} (top) displays 13 of the 25 reconstructed spectral bands of view(2,2).
The band-wise visualizations exhibit a well-defined structure, clear textural details, and minimal artifacts. Both the pseudo-RGB visualization and spectral reconstruction results demonstrate the reliability and practical effectiveness of ADLIS under real-world imaging conditions. 

For quantitative evaluation of reconstruction fidelity, we analyze spectral signatures at three spatial points from the central view, with reference data collected using a probe spectrometer (Malvern Panalytical ASD TerraSpec 4 Hi-Res Mineral Spectrometer). The predicted spectral curves show close agreement with the spectrometer at all marker points in Fig. \ref{fig:real} (top).
For an inspection across the angular dimensions, Fig. \ref{fig:real} (bottom) shows the reconstruction results of the four band-wise light fields. Unlike the MLA-based solutions, which suffer greatly from the vignetting effect \cite{xiong2017snapshot}, where the central view has higher intensity and higher signal-to-noise ratio than the corner views, our method achieves a band-wise light field with a uniform intensity distribution regardless of the views as in the simulation and decently recovers the 5D-SLF, although some noise is visible owing to the imperfect measurements and some inevitable gap between the simulation and real-world. 
To further validate the angular reconstruction reliability, we estimate the disparity map for the ROI. The results in Fig. \ref{fig:real} (top) also show the detailed relative depth variations across the scene surface, thus confirming the robust angular reconstruction capability of the proposed ADLIS.


\section{DISCUSSION AND LIMITATION}
Despite sprouting in many front visual applications, high-dimensional imaging systems share the same unavoidable mutual restraint between the resolutions of different dimensions, especially when it comes to a miniaturized system whose encoding capability is limited.
Unfortunately, based on the principle of physical light splitting, existing single-detector systems for multi-dimensional light-field imaging allocate a portion of the spatial sensor pixels to record information from other dimensions, leading to under-utilization of the detector and limitation of spatial resolution.

\begin{table}[!htb]
\centering
\caption{Comparison of Hyperspectral Light Field Imagers in \textbf{S}patial \textbf{R}esolution, \textbf{A}ngular \textbf{R}esolution, \textbf{Total} number of sensor \textbf{Pixels} and the \textbf{SIE} which is defined by $\frac{\mathbf{SR}}{\mathbf{Total Pixels}}$.}
\renewcommand{\arraystretch}{1.2}  
\setlength{\tabcolsep}{2pt}  
\resizebox{80mm}{!}{
\begin{tabular}{@{}ccccc@{}}  
\hline
\rowcolor[HTML]{FFFFFF} 
& \textbf{SR} 
& \textbf{AR} 
& \textbf{Total Pixels} 
& \textbf{SIE} \\ \hline
Light field IMS \cite{cui2020snapshot}                                             
& 180 × 180                   
& 18 × 13                     
& 3296 × 2472                            
& 0.40\%        \\
$C^3$SLFI\cite{marquez2020compressive}      
& 761 × 789                   
& 4 × 4                       
& 1032 × 776                             
& 75\%         \\
Hyper-LIFT \cite{cui2021snapshot}                                                         
& 270 × 270                   
& 4 × 4                       
& 4864 × 3232                            
& 0.46\%         \\
SLIM \cite{hua2022ultra}          
& 75 × 75                     
& 48 × 48                     
& 5472 × 3648                            
& 0.03\%         \\
Inkjet-printed-based \cite{zhang2024compact}
& 96 × 96                     
& 9 × 9                       
& 2048 × 2048                            
& 0.22\%         \\
ADLIS (ours)                                                   
& 2448 × 2048                 
& 3 × 3                       
& 2448 × 2048                            
& \textbf{100\%}     \\ \hline
\end{tabular}}
\label{tab5}
\end{table}

\begin{figure*}[!htb]
\centering
\includegraphics[width=1\linewidth]{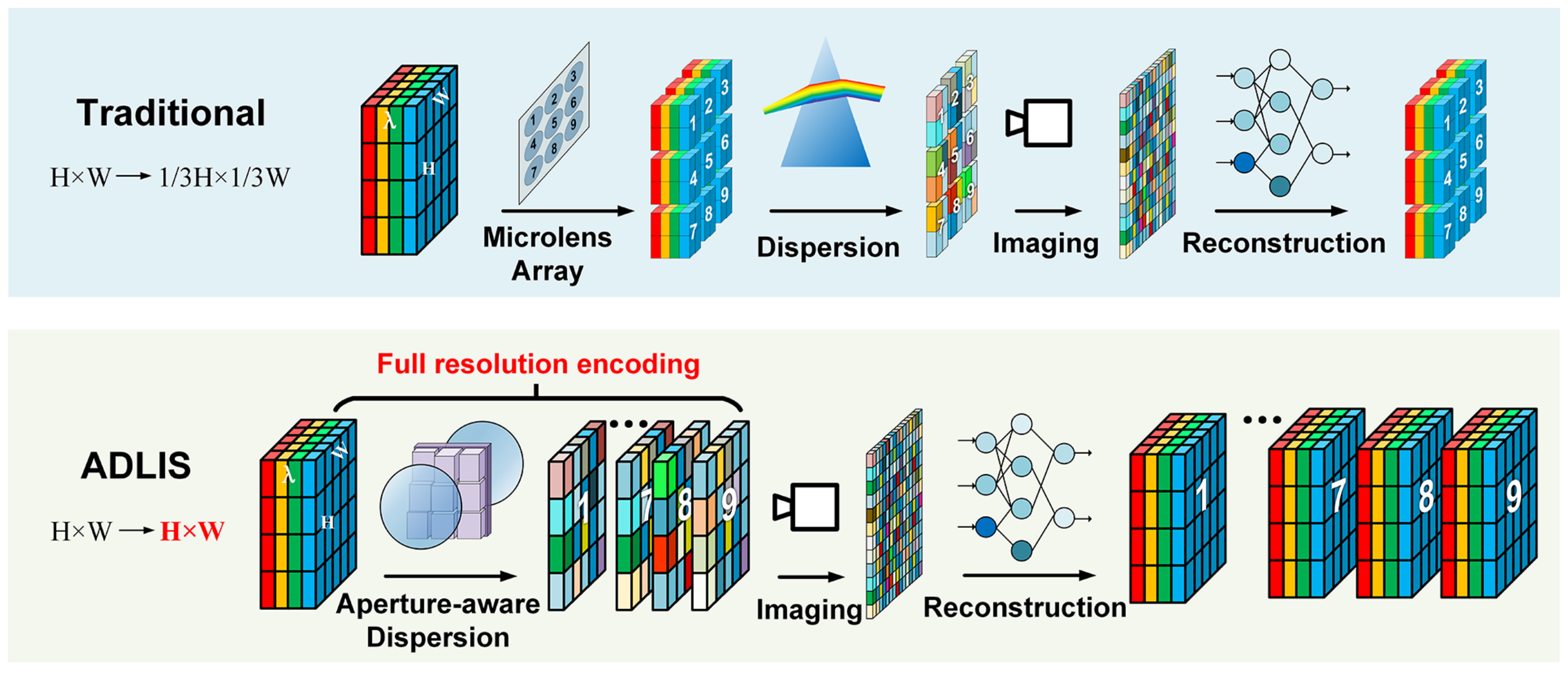}
  \caption{ We compare the inherent difference between the traditional MLA-based solutions and the proposed ADLIS. There is a secondary deflection caused by the MLA restricts each sensor pixel to receiving light only from a specific subset of angles. In contrast, ADLIS captures the full-resolution SLF directly at the aperture. Consequently, the intensity at each pixel constitutes a weighted sum of contributions from all viewpoints. }
  \label{fig:compare}
\end{figure*}

\begin{table}[!htb]
\centering
\caption{Noise analysis.}
\renewcommand{\arraystretch}{1.2}  
\setlength{\tabcolsep}{5pt}  
\resizebox{80mm}{!}{
\begin{tabular}{cc|cc} 
\hline
\textbf{Encoder}       & \textbf{Algorithm} & \textbf{PSNR}  & ($+\sigma=0.01$)      \\ \hline
CFA           & Restormer\cite{zamir2022restormer} & 35.07 & 33.39          \\
ADLIS-1frame  & Restormer\cite{zamir2022restormer} & 38.65 & 37.63          \\
ADLIS-2frames & Restormer\cite{zamir2022restormer} & 41.36 & \textbf{40.55} \\ \hline
CFA           & EDSR\cite{lim2017enhanced}      & 32.95 & 30.41          \\
ADLIS-1frame  & EDSR\cite{lim2017enhanced}       & 36.28 & 34.45          \\
ADLIS-2frames & EDSR\cite{lim2017enhanced}       & 38.22 & \textbf{36.74} \\ \hline
CFA           & U-Net\cite{ronneberger2015u}     & 30.11 & 27.08          \\
ADLIS-1frame  & U-Net\cite{ronneberger2015u}       & 31.25 & 29.82          \\
ADLIS-2frames & U-Net\cite{ronneberger2015u}       & 35.09 & \textbf{33.54} \\ \hline
\end{tabular}}
\label{tab6}
\end{table}

To enable a more focused analysis on spatial performance of imaging systems in high-dimensional light field acquisition, we define the Spatial Information Efficiency (SIE) metric as the ratio between the spatial resolution of the reconstructed data cube and the total number of sensor pixels.
In Table \ref{tab5}, ADLIS demonstrates a superior spatial SIE compared with other state-of-the-art single-detector 5D-SLF imagers \cite{cui2020snapshot,marquez2020compressive,zhang2024compact,hua2022ultra,cui2021snapshot}. 
It is because that for ADLIS, information from all viewpoints is both encoded and recorded in the full-spatial resolution, rather than being mapped to distinct pixel regions for different sub-views, as is shown in Fig. \ref{fig:compare}. Supported by a novel, high-quality real-world SLF dataset that provides physical constraints for computational reconstruction, our ADLIS enables high-dimensional imaging while maintaining full-spatial resolution, indicating the advantage in sensor utilization efficiency.
Additionally, measurements are often degraded by various noise sources in real-data experiments, causing the acquired data to deviate from the ideal forward model. To evaluate the impact of noise on reconstruction performance across different encoding models and algorithms, we introduce Gaussian noise into representative ADLIS for a noise study. The experimental results, summarized in Table. \ref{tab6}, reveals that whatever the decoding algorithm, the proposed ADLIS achieves a better system robustness and algorithmic stability under the same noise.

Despite the fact that the current angular resolution is limited by the 3 × 3 aperture configuration, it can be enhanced by re-optimizing the phase plate design to support larger viewpoint arrays.
Notably, the RealSLF dataset supports the E2E optimization of SLF encoders with up to 7 × 5 viewpoints, and quartz crystal fabrication remains considerably more cost-effective than complex optical alternatives. Thus, ADLIS presents a viable pathway toward high-resolution high-dimensional light field acquisition with favorable cost efficiency in practical applications.

\section{CONCLUSION}

In conclusion, we demonstrate the ADLIS, a miniaturized, cost-effective high-dimensional imaging system with a compact encoding and joint optimization framework, which is able to record the 5D-SLF at full spatial resolution.
In this work, we leverage the phase retardation characteristics of birefringent crystals to elegantly achieve aperture-aware dispersion. 
The encoding capacity of the aperture can be further enhanced through multi-frame acquisition schemes, whereas the optical design can be jointly optimized with the reconstruction network within an E2E framework.
The comprehensive experimental results validate the significant advantages of ADLIS in high-dimensional light field modulation and imaging performance.


The proposed ADLIS, with its co-optimized spectro-angular encoding aperture, provides practical solutions for applications requiring high-dimensional visual perception, including embodied intelligence systems requiring material-aware interaction, robotic vision demanding precise geometry-spectrum understanding, and autonomous systems operating under extreme dynamic range conditions.
By integrating interpretable optical encoding with data-driven reconstruction, ADLIS establishes a new paradigm for high-dimensional imaging where physics-based models and AI algorithms are complementarily fused to achieve enhanced practical performance. This approach provides valuable inspiration for the development of future compact high-resolution imaging systems.
Furthermore, it should be noted that the proposed ADLIS can be extended to high-resolution imaging and surface recovery of objects in extreme high-dynamic-range (HDR) environments due to the information preservation capability of polarized multi-frame measurements under overexposed or underexposed conditions.
Looking forward, we plan to extend the perceptual dimensions of the ADLIS to include polarization sensitivity, with the aim of developing a compact plenoptic imaging system capable of addressing broader high-dimensional visual challenges.

\bibliographystyle{IEEEtran}
\bibliography{ADLIS}

\begin{thebibliography}{10}
\providecommand{\url}[1]{#1}
\csname url@samestyle\endcsname
\providecommand{\newblock}{\relax}
\providecommand{\bibinfo}[2]{#2}
\providecommand{\BIBentrySTDinterwordspacing}{\spaceskip=0pt\relax}
\providecommand{\BIBentryALTinterwordstretchfactor}{4}
\providecommand{\BIBentryALTinterwordspacing}{\spaceskip=\fontdimen2\font plus
\BIBentryALTinterwordstretchfactor\fontdimen3\font minus
  \fontdimen4\font\relax}
\providecommand{\BIBforeignlanguage}[2]{{%
\expandafter\ifx\csname l@#1\endcsname\relax
\typeout{** WARNING: IEEEtran.bst: No hyphenation pattern has been}%
\typeout{** loaded for the language `#1'. Using the pattern for}%
\typeout{** the default language instead.}%
\else
\language=\csname l@#1\endcsname
\fi
#2}}
\providecommand{\BIBdecl}{\relax}
\BIBdecl

\bibitem{ziegler2017acquisition}
M.~Ziegler, R.~op~het Veld, J.~Keinert, and F.~Zilly, ``Acquisition system for
  dense lightfield of large scenes,'' in \emph{2017 3DTV Conference: the true
  vision-capture, transmission and display of 3D Video (3DTV-CON)}.\hskip 1em
  plus 0.5em minus 0.4em\relax IEEE, 2017, pp. 1--4.

\bibitem{stanford2008lightfield}
\BIBentryALTinterwordspacing
Stanford, ``Stanford light field archive,'' 2008, accessed: 2024-10-04.
  [Online]. Available: \url{http://lightfield.stanford.edu/}
\BIBentrySTDinterwordspacing

\bibitem{ng2005light}
R.~Ng, M.~Levoy, M.~Br{\'e}dif, G.~Duval, M.~Horowitz, and P.~Hanrahan, ``Light
  field photography with a hand-held plenoptic camera,'' Ph.D. dissertation,
  Stanford university, 2005.

\bibitem{wu2017light}
G.~Wu, B.~Masia, A.~Jarabo, Y.~Zhang, L.~Wang, Q.~Dai, T.~Chai, and Y.~Liu,
  ``Light field image processing: An overview,'' \emph{IEEE Journal of Selected
  Topics in Signal Processing}, vol.~11, no.~7, pp. 926--954, 2017.

\bibitem{wagadarikar2008single}
A.~Wagadarikar, R.~John, R.~Willett, and D.~Brady, ``Single disperser design
  for coded aperture snapshot spectral imaging,'' \emph{Applied optics},
  vol.~47, no.~10, pp. B44--B51, 2008.

\bibitem{arce2013compressive}
G.~R. Arce, D.~J. Brady, L.~Carin, H.~Arguello, and D.~S. Kittle, ``Compressive
  coded aperture spectral imaging: An introduction,'' \emph{IEEE Signal
  Processing Magazine}, vol.~31, no.~1, pp. 105--115, 2013.

\bibitem{cao2011prism}
X.~Cao, H.~Du, X.~Tong, Q.~Dai, and S.~Lin, ``A prism-mask system for
  multispectral video acquisition,'' \emph{IEEE transactions on pattern
  analysis and machine intelligence}, vol.~33, no.~12, pp. 2423--2435, 2011.

\bibitem{cao2016computational}
X.~Cao, T.~Yue, X.~Lin, S.~Lin, X.~Yuan, Q.~Dai, L.~Carin, and D.~J. Brady,
  ``Computational snapshot multispectral cameras: Toward dynamic capture of the
  spectral world,'' \emph{IEEE Signal Processing Magazine}, vol.~33, no.~5, pp.
  95--108, 2016.

\bibitem{shi2025prior}
Z.~Shi, Z.~Xu, L.~Cai, H.~Ye, L.~Chen, Q.~Shen, and X.~Cao, ``Prior image
  guided snapshot high-resolution spectral imaging in near infrared,''
  \emph{IEEE Transactions on Image Processing}, 2025.

\bibitem{deng2025compact}
Z.~Deng, Z.~Shi, C.~Huang, S.~Li, and X.~Cao, ``Compact snapshot
  multispectral-depth imaging system with shared-modal multi-bandpass lenslet
  array (smla),'' \emph{Optics Letters}, vol.~50, no.~11, pp. 3568--3571, 2025.

\bibitem{deng2025shared}
Z.~Deng, Z.~Shi, S.~Li, C.~Huang, and X.~Cao, ``Shared-modality multi-bandpass
  filtering for plug-and-play multispectral depth imaging,'' \emph{Optics
  Express}, vol.~33, no.~14, pp. 29\,895--29\,911, 2025.

\bibitem{chen2023notch}
L.~Chen, L.~Cai, E.~Huang, Y.~Zhou, T.~Yue, and X.~Cao, ``A notch-mask and
  dual-prism system for snapshot spectral imaging,'' \emph{Optics and Lasers in
  Engineering}, vol. 165, p. 107544, 2023.

\bibitem{gehrig2024low}
D.~Gehrig and D.~Scaramuzza, ``Low-latency automotive vision with event
  cameras,'' \emph{Nature}, vol. 629, no. 8014, pp. 1034--1040, 2024.

\bibitem{almalioglu2022deep}
Y.~Almalioglu, M.~Turan, N.~Trigoni, and A.~Markham, ``Deep learning-based
  robust positioning for all-weather autonomous driving,'' \emph{Nature machine
  intelligence}, vol.~4, no.~9, pp. 749--760, 2022.

\bibitem{zhao2017heterogeneous}
Y.~Zhao, T.~Yue, L.~Chen, H.~Wang, Z.~Ma, D.~J. Brady, and X.~Cao,
  ``Heterogeneous camera array for multispectral light field imaging,''
  \emph{Optics Express}, vol.~25, no.~13, pp. 14\,008--14\,022, 2017.

\bibitem{jakubovic2018high}
R.~Jakubovic, D.~Guha, S.~Gupta, M.~Lu, J.~Jivraj, B.~A. Standish, M.~K. Leung,
  A.~Mariampillai, K.~Lee, P.~Siegler \emph{et~al.}, ``High speed, high density
  intraoperative 3d optical topographical imaging with efficient registration
  to mri and ct for craniospinal surgical navigation,'' \emph{Scientific
  reports}, vol.~8, no.~1, p. 14894, 2018.

\bibitem{kok2020accurate}
E.~N. Kok, R.~Eppenga, K.~F. Kuhlmann, H.~C. Groen, R.~van Veen, J.~M. van
  Dieren, T.~R. de~Wijkerslooth, M.~van Leerdam, D.~M. Lambregts, W.~J. Heerink
  \emph{et~al.}, ``Accurate surgical navigation with real-time tumor tracking
  in cancer surgery,'' \emph{NPJ precision oncology}, vol.~4, no.~1, p.~8,
  2020.

\bibitem{matinfar2023sonification}
S.~Matinfar, M.~Salehi, D.~Suter, M.~Seibold, S.~Dehghani, N.~Navab,
  F.~Wanivenhaus, P.~F{\"u}rnstahl, M.~Farshad, and N.~Navab, ``Sonification as
  a reliable alternative to conventional visual surgical navigation,''
  \emph{Scientific Reports}, vol.~13, no.~1, p. 5930, 2023.

\bibitem{liodakis2022polarized}
I.~Liodakis, A.~P. Marscher, I.~Agudo, A.~V. Berdyugin, M.~I. Bernardos,
  G.~Bonnoli, G.~A. Borman, C.~Casadio, V.~Casanova, E.~Cavazzuti
  \emph{et~al.}, ``Polarized blazar x-rays imply particle acceleration in
  shocks,'' \emph{Nature}, vol. 611, no. 7937, pp. 677--681, 2022.

\bibitem{roberts2021rapid}
O.~Roberts, P.~Veres, M.~Baring, M.~Briggs, C.~Kouveliotou, E.~Bissaldi,
  G.~Younes, S.~Chastain, J.~DeLaunay, D.~Huppenkothen \emph{et~al.}, ``Rapid
  spectral variability of a giant flare from a magnetar in ngc 253,''
  \emph{Nature}, vol. 589, no. 7841, pp. 207--210, 2021.

\bibitem{dai2023coastal}
Y.~Dai, S.~Yang, D.~Zhao, C.~Hu, W.~Xu, D.~M. Anderson, Y.~Li, X.-P. Song,
  D.~G. Boyce, L.~Gibson \emph{et~al.}, ``Coastal phytoplankton blooms expand
  and intensify in the 21st century,'' \emph{Nature}, vol. 615, no. 7951, pp.
  280--284, 2023.

\bibitem{basedow1995hydice}
R.~W. Basedow, D.~C. Carmer, and M.~E. Anderson, ``Hydice system:
  Implementation and performance,'' in \emph{Imaging Spectrometry}, vol.
  2480.\hskip 1em plus 0.5em minus 0.4em\relax SPIE, 1995, pp. 258--267.

\bibitem{hsu2017line}
Y.~J. Hsu, C.-C. Chen, C.-H. Huang, C.-H. Yeh, L.-Y. Liu, and S.-Y. Chen,
  ``Line-scanning hyperspectral imaging based on structured illumination
  optical sectioning,'' \emph{Biomedical Optics Express}, vol.~8, no.~6, pp.
  3005--3016, 2017.

\bibitem{abdo2019spatial}
M.~Abdo, V.~Badilita, and J.~Korvink, ``Spatial scanning hyperspectral imaging
  combining a rotating slit with a dove prism,'' \emph{Optics Express},
  vol.~27, no.~15, pp. 20\,290--20\,304, 2019.

\bibitem{zhu2018hyperspectral}
K.~Zhu, Y.~Xue, Q.~Fu, S.~B. Kang, X.~Chen, and J.~Yu, ``Hyperspectral light
  field stereo matching,'' \emph{IEEE transactions on pattern analysis and
  machine intelligence}, vol.~41, no.~5, pp. 1131--1143, 2018.

\bibitem{holloway2014generalized}
J.~Holloway, K.~Mitra, S.~J. Koppal, and A.~N. Veeraraghavan, ``Generalized
  assorted camera arrays: Robust cross-channel registration and applications,''
  \emph{IEEE Transactions on Image Processing}, vol.~24, no.~3, pp. 823--835,
  2014.

\bibitem{xiong2017snapshot}
Z.~Xiong, L.~Wang, H.~Li, D.~Liu, and F.~Wu, ``Snapshot hyperspectral light
  field imaging,'' in \emph{Proceedings of the IEEE Conference on Computer
  Vision and Pattern Recognition}, 2017, pp. 3270--3278.

\bibitem{zhu2018complete}
S.~Zhu, L.~Gao, Y.~Zhang, J.~Lin, and P.~Jin, ``Complete plenoptic imaging
  using a single detector,'' \emph{Optics Express}, vol.~26, no.~20, pp.
  26\,495--26\,510, 2018.

\bibitem{marquez2019snapshot}
M.~Marquez, H.~Rueda, E.~Vera, and H.~Arguello, ``Snapshot compressive spectral
  light field tensor imaging,'' in \emph{Computational Optical Sensing and
  Imaging}.\hskip 1em plus 0.5em minus 0.4em\relax Optica Publishing Group,
  2019, pp. CTu2A--6.

\bibitem{lv2020snapshot}
X.~Lv, Y.~Li, S.~Zhu, X.~Guo, J.~Zhang, J.~Lin, and P.~Jin, ``Snapshot spectral
  polarimetric light field imaging using a single detector,'' \emph{Optics
  letters}, vol.~45, no.~23, pp. 6522--6525, 2020.

\bibitem{cui2020snapshot}
Q.~Cui, J.~Park, R.~Theodore~Smith, and L.~Gao, ``Snapshot hyperspectral light
  field imaging using image mapping spectrometry,'' \emph{Optics letters},
  vol.~45, no.~3, pp. 772--775, 2020.

\bibitem{marquez2020compressive}
M.~Marquez, H.~Rueda-Chacon, and H.~Arguello, ``Compressive spectral light
  field image reconstruction via online tensor representation,'' \emph{IEEE
  Transactions on Image Processing}, vol.~29, pp. 3558--3568, 2020.

\bibitem{zhang2024compact}
Q.~Zhang, M.~Schambach, Q.~Jin, M.~Heizmann, and U.~Lemmer, ``Compact
  multispectral light field camera based on an inkjet-printed microlens array
  and color filter array,'' \emph{Optics Express}, vol.~32, no.~13, pp.
  23\,510--23\,523, 2024.

\bibitem{xue2017catadioptric}
Y.~Xue, K.~Zhu, Q.~Fu, X.~Chen, and J.~Yu, ``Catadioptric hyperspectral light
  field imaging,'' in \emph{Proceedings of the IEEE International Conference on
  Computer Vision}, 2017, pp. 985--993.

\bibitem{hua2022ultra}
X.~Hua, Y.~Wang, S.~Wang, X.~Zou, Y.~Zhou, L.~Li, F.~Yan, X.~Cao, S.~Xiao,
  D.~P. Tsai \emph{et~al.}, ``Ultra-compact snapshot spectral light-field
  imaging,'' \emph{Nature communications}, vol.~13, no.~1, p. 2732, 2022.

\bibitem{cui2021snapshot}
Q.~Cui, J.~Park, Y.~Ma, and L.~Gao, ``Snapshot hyperspectral light field
  tomography,'' \emph{Optica}, vol.~8, no.~12, pp. 1552--1558, 2021.

\bibitem{zhao2023coded}
R.~Zhao, Q.~Cui, Z.~Wang, and L.~Gao, ``Coded aperture snapshot hyperspectral
  light field tomography,'' \emph{Optics Express}, vol.~31, no.~22, pp.
  37\,336--37\,347, 2023.

\bibitem{li2025realslf}
S.~Li, T.~Lv, C.~Huang, H.~Ye, Z.~Deng, L.~Hu, Q.~Li, C.~Zi, L.~Chen, and
  X.~Cao, ``Realslf and flexidim: towards practical spectral light field
  imaging,'' \emph{Optics Express}, vol.~33, no.~21, pp. 45\,049--45\,065,
  2025.

\bibitem{ghanekar2024passive}
B.~Ghanekar, S.~S. Khan, P.~Sharma, S.~Singh, V.~Boominathan, K.~Mitra, and
  A.~Veeraraghavan, ``Passive snapshot coded aperture dual-pixel rgb-d
  imaging,'' in \emph{Proceedings of the IEEE/CVF Conference on Computer Vision
  and Pattern Recognition}, 2024, pp. 25\,348--25\,357.

\bibitem{meng2020gap}
Z.~Meng, S.~Jalali, and X.~Yuan, ``Gap-net for snapshot compressive imaging,''
  \emph{arXiv preprint arXiv:2012.08364}, 2020.

\bibitem{qiao2020deep}
M.~Qiao, Z.~Meng, J.~Ma, and X.~Yuan, ``Deep learning for video compressive
  sensing,'' \emph{Apl Photonics}, vol.~5, no.~3, 2020.

\bibitem{fan2024dispersion}
Y.~Fan, W.~Huang, F.~Zhu, X.~Liu, C.~Jin, C.~Guo, Y.~An, Y.~Kivshar, C.-W. Qiu,
  and W.~Li, ``Dispersion-assisted high-dimensional photodetector,''
  \emph{Nature}, pp. 1--7, 2024.

\bibitem{fu2025miniaturized}
B.~Fu, X.~Zhou, T.~Li, H.~Zhu, Z.~Liu, S.~Zheng, Y.~Zhou, Y.~Yu, X.~Cao,
  S.~Wang \emph{et~al.}, ``Miniaturized high-efficiency snapshot polarimetric
  stereoscopic imaging,'' \emph{Optica}, vol.~12, no.~3, pp. 391--398, 2025.

\bibitem{lopez2024depth}
J.~Lopez, E.~Vargas, and H.~Arguello, ``Depth estimation from a single optical
  encoded image using a learned colored-coded aperture,'' \emph{IEEE
  Transactions on Computational Imaging}, vol.~10, pp. 752--761, 2024.

\bibitem{lv2023aperture}
T.~Lv, H.~Ye, Q.~Yuan, Z.~Shi, Y.~Wang, S.~Wang, and X.~Cao, ``Aperture
  diffraction for compact snapshot spectral imaging,'' in \emph{Proceedings of
  the IEEE/CVF International Conference on Computer Vision}, 2023, pp.
  10\,574--10\,584.

\bibitem{shi2023compact}
Z.~Shi, H.~Ye, T.~Lv, Y.~Wang, and X.~Cao, ``Compact self-adaptive coding for
  spectral compressive sensing,'' in \emph{2023 IEEE International Conference
  on Computational Photography (ICCP)}.\hskip 1em plus 0.5em minus 0.4em\relax
  IEEE, 2023, pp. 1--12.

\bibitem{lv2024efficient}
T.~Lv, L.~Hu, S.~Li, C.~Huang, and X.~Cao, ``Efficient snapshot spectral
  imaging: Calibration-free parallel structure with aperture diffraction
  fusion,'' in \emph{European Conference on Computer Vision}.\hskip 1em plus
  0.5em minus 0.4em\relax Springer, 2024, pp. 93--110.

\bibitem{cai2024exploring}
L.~Cai, X.~Dong, K.~Zhou, and X.~Cao, ``Exploring video denoising in thermal
  infrared imaging: Physics-inspired noise generator, dataset, and model,''
  \emph{IEEE Transactions on Image Processing}, vol.~33, pp. 3839--3854, 2024.

\bibitem{sitzmann2018end}
V.~Sitzmann, S.~Diamond, Y.~Peng, X.~Dun, S.~Boyd, W.~Heidrich, F.~Heide, and
  G.~Wetzstein, ``End-to-end optimization of optics and image processing for
  achromatic extended depth of field and super-resolution imaging,'' \emph{ACM
  Transactions on Graphics (TOG)}, vol.~37, no.~4, pp. 1--13, 2018.

\bibitem{chang2019deep}
J.~Chang and G.~Wetzstein, ``Deep optics for monocular depth estimation and 3d
  object detection,'' in \emph{Proceedings of the IEEE/CVF International
  Conference on Computer Vision}, 2019, pp. 10\,193--10\,202.

\bibitem{ikoma2021depth}
H.~Ikoma, C.~M. Nguyen, C.~A. Metzler, Y.~Peng, and G.~Wetzstein, ``Depth from
  defocus with learned optics for imaging and occlusion-aware depth
  estimation,'' in \emph{2021 IEEE International Conference on Computational
  Photography (ICCP)}.\hskip 1em plus 0.5em minus 0.4em\relax IEEE, 2021, pp.
  1--12.

\bibitem{shi2024split}
Z.~Shi, I.~Chugunov, M.~Bijelic, G.~C{\^o}t{\'e}, J.~Yeom, Q.~Fu, H.~Amata,
  W.~Heidrich, and F.~Heide, ``Split-aperture 2-in-1 computational cameras,''
  \emph{ACM Transactions on Graphics (TOG)}, vol.~43, no.~4, pp. 1--19, 2024.

\bibitem{zhang2022herosnet}
X.~Zhang, Y.~Zhang, R.~Xiong, Q.~Sun, and J.~Zhang, ``Herosnet: Hyperspectral
  explicable reconstruction and optimal sampling deep network for snapshot
  compressive imaging,'' in \emph{Proceedings of the IEEE/CVF Conference on
  Computer Vision and Pattern Recognition}, 2022, pp. 17\,532--17\,541.

\bibitem{shi2018hscnn+}
Z.~Shi, C.~Chen, Z.~Xiong, D.~Liu, and F.~Wu, ``Hscnn+: Advanced cnn-based
  hyperspectral recovery from rgb images,'' in \emph{Proceedings of the IEEE
  Conference on Computer Vision and Pattern Recognition Workshops}, 2018, pp.
  939--947.

\bibitem{li2022quantization}
L.~Li, L.~Wang, W.~Song, L.~Zhang, Z.~Xiong, and H.~Huang, ``Quantization-aware
  deep optics for diffractive snapshot hyperspectral imaging,'' in
  \emph{Proceedings of the IEEE/CVF Conference on Computer Vision and Pattern
  Recognition}, 2022, pp. 19\,780--19\,789.

\bibitem{hershko2019multicolor}
E.~Hershko, L.~E. Weiss, T.~Michaeli, and Y.~Shechtman, ``Multicolor
  localization microscopy and point-spread-function engineering by deep
  learning,'' \emph{Optics express}, vol.~27, no.~5, pp. 6158--6183, 2019.

\bibitem{horstmeyer2017convolutional}
R.~Horstmeyer, R.~Y. Chen, B.~Kappes, and B.~Judkewitz, ``Convolutional neural
  networks that teach microscopes how to image,'' \emph{arXiv preprint
  arXiv:1709.07223}, 2017.

\bibitem{nehme2020deepstorm3d}
E.~Nehme, D.~Freedman, R.~Gordon, B.~Ferdman, L.~E. Weiss, O.~Alalouf, T.~Naor,
  R.~Orange, T.~Michaeli, and Y.~Shechtman, ``Deepstorm3d: dense 3d
  localization microscopy and psf design by deep learning,'' \emph{Nature
  methods}, vol.~17, no.~7, pp. 734--740, 2020.

\bibitem{marco2017deeptof}
J.~Marco, Q.~Hernandez, A.~Munoz, Y.~Dong, A.~Jarabo, M.~H. Kim, X.~Tong, and
  D.~Gutierrez, ``Deeptof: off-the-shelf real-time correction of multipath
  interference in time-of-flight imaging,'' \emph{ACM Transactions on Graphics
  (ToG)}, vol.~36, no.~6, pp. 1--12, 2017.

\bibitem{su2018deep}
S.~Su, F.~Heide, G.~Wetzstein, and W.~Heidrich, ``Deep end-to-end
  time-of-flight imaging,'' in \emph{Proceedings of the IEEE Conference on
  Computer Vision and Pattern Recognition}, 2018, pp. 6383--6392.

\bibitem{turpin2018light}
A.~Turpin, I.~Vishniakou, and J.~D. Seelig, ``Light scattering control with
  neural networks in transmission and reflection,'' \emph{Arxiv: 180505602
  [Cs]}, 2018.

\bibitem{yang2024end}
X.~Yang, M.~Souza, K.~Wang, P.~Chakravarthula, Q.~Fu, and W.~Heidrich,
  ``End-to-end hybrid refractive-diffractive lens design with differentiable
  ray-wave model,'' in \emph{SIGGRAPH Asia 2024 Conference Papers}, 2024, pp.
  1--11.

\bibitem{zhang2022end}
B.~Zhang, X.~Yuan, C.~Deng, Z.~Zhang, J.~Suo, and Q.~Dai, ``End-to-end snapshot
  compressed super-resolution imaging with deep optics,'' \emph{Optica},
  vol.~9, no.~4, pp. 451--454, 2022.

\bibitem{baek2021single}
S.-H. Baek, H.~Ikoma, D.~S. Jeon, Y.~Li, W.~Heidrich, G.~Wetzstein, and M.~H.
  Kim, ``Single-shot hyperspectral-depth imaging with learned diffractive
  optics,'' in \emph{Proceedings of the IEEE/CVF International Conference on
  Computer Vision}, 2021, pp. 2651--2660.

\bibitem{zheng2023close}
C.~Zheng, G.~Zhao, and P.~So, ``Close the design-to-manufacturing gap in
  computational optics with a'real2sim'learned two-photon neural lithography
  simulator,'' in \emph{SIGGRAPH Asia 2023 Conference Papers}, 2023, pp. 1--9.

\bibitem{zamir2022restormer}
S.~W. Zamir, A.~Arora, S.~Khan, M.~Hayat, F.~S. Khan, and M.-H. Yang,
  ``Restormer: Efficient transformer for high-resolution image restoration,''
  in \emph{Proceedings of the IEEE/CVF conference on computer vision and
  pattern recognition}, 2022, pp. 5728--5739.

\bibitem{smith1931cie}
T.~Smith and J.~Guild, ``The cie colorimetric standards and their use,''
  \emph{Transactions of the optical society}, vol.~33, no.~3, p.~73, 1931.

\bibitem{kruse1993spectral}
F.~A. Kruse, A.~B. Lefkoff, J.~W. Boardman, K.~B. Heidebrecht, A.~Shapiro,
  P.~Barloon, and A.~F. Goetz, ``The spectral image processing system
  (sips)—interactive visualization and analysis of imaging spectrometer
  data,'' \emph{Remote sensing of environment}, vol.~44, no. 2-3, pp. 145--163,
  1993.

\bibitem{monakhova2020spectral}
K.~Monakhova, K.~Yanny, N.~Aggarwal, and L.~Waller, ``Spectral diffusercam:
  lensless snapshot hyperspectral imaging with a spectral filter array,''
  \emph{Optica}, vol.~7, no.~10, pp. 1298--1307, 2020.

\bibitem{ipol.2013.26}
J.~Sánchez~Pérez, E.~Meinhardt-Llopis, and G.~Facciolo, ``{TV-L1 Optical Flow
  Estimation},'' \emph{{Image Processing On Line}}, vol.~3, pp. 137--150, 2013,
  \url{https://doi.org/10.5201/ipol.2013.26}.

\bibitem{meng2023deep}
Z.~Meng, X.~Yuan, and S.~Jalali, ``Deep unfolding for snapshot compressive
  imaging,'' \emph{International Journal of Computer Vision}, vol. 131, no.~11,
  pp. 2933--2958, 2023.

\bibitem{ronneberger2015u}
O.~Ronneberger, P.~Fischer, and T.~Brox, ``U-net: Convolutional networks for
  biomedical image segmentation,'' in \emph{International Conference on Medical
  image computing and computer-assisted intervention}.\hskip 1em plus 0.5em
  minus 0.4em\relax Springer, 2015, pp. 234--241.

\bibitem{zhao2020hierarchical}
Y.~Zhao, L.-M. Po, Q.~Yan, W.~Liu, and T.~Lin, ``Hierarchical regression
  network for spectral reconstruction from rgb images,'' in \emph{Proceedings
  of the IEEE/CVF Conference on Computer Vision and Pattern Recognition
  Workshops}, 2020, pp. 422--423.

\bibitem{wang2025sˆ}
J.~Wang, K.~Li, Y.~Zhang, X.~Yuan, and Z.~Tao, ``Sˆ 2-transformer for
  mask-aware hyperspectral image reconstruction,'' \emph{IEEE Transactions on
  Pattern Analysis and Machine Intelligence}, 2025.

\bibitem{hu2022hdnet}
X.~Hu, Y.~Cai, J.~Lin, H.~Wang, X.~Yuan, Y.~Zhang, R.~Timofte, and L.~Van~Gool,
  ``Hdnet: High-resolution dual-domain learning for spectral compressive
  imaging,'' in \emph{Proceedings of the IEEE/CVF Conference on Computer Vision
  and Pattern Recognition}, 2022, pp. 17\,542--17\,551.

\bibitem{lim2017enhanced}
B.~Lim, S.~Son, H.~Kim, S.~Nah, and K.~Mu~Lee, ``Enhanced deep residual
  networks for single image super-resolution,'' in \emph{Proceedings of the
  IEEE conference on computer vision and pattern recognition workshops}, 2017,
  pp. 136--144.

\bibitem{zamir2020learning}
S.~W. Zamir, A.~Arora, S.~Khan, M.~Hayat, F.~S. Khan, M.-H. Yang, and L.~Shao,
  ``Learning enriched features for real image restoration and enhancement,'' in
  \emph{European conference on computer vision}.\hskip 1em plus 0.5em minus
  0.4em\relax Springer, 2020, pp. 492--511.

\bibitem{zamir2021multi}
S.~W. Zamir, A.~Arora, S.~Khan, M.~Hayat, F.~S. Khan, M.~Yang, and L.~Shao,
  ``Multi-stage progressive image restoration,'' in \emph{Proceedings of the
  IEEE/CVF conference on computer vision and pattern recognition}, 2021, pp.
  14\,821--14\,831.

\end{thebibliography}

\section*{Acknowledgment}
This research was supported by the National Key Research and Development Program of China (2023YFF0713300). 
The authors would like to thank the NJU-ZPTECH University-Enterprise Joint Laboratory for Computational Spectral Imaging for providing hardware support. The authors also thank Dr. Zhan Shi, Dr. Shiqiao Li, Dr. Lijing Cai and Dr. Zhiwei Deng for usefull suggestions.
Selected optical components were fabricated with the support of the Fuzhou Optoforon Optoelectronics Co., Ltd.

\section{Biography Section}
\begin{IEEEbiography}[{\includegraphics[width=1in,height=1.25in,clip,keepaspectratio]{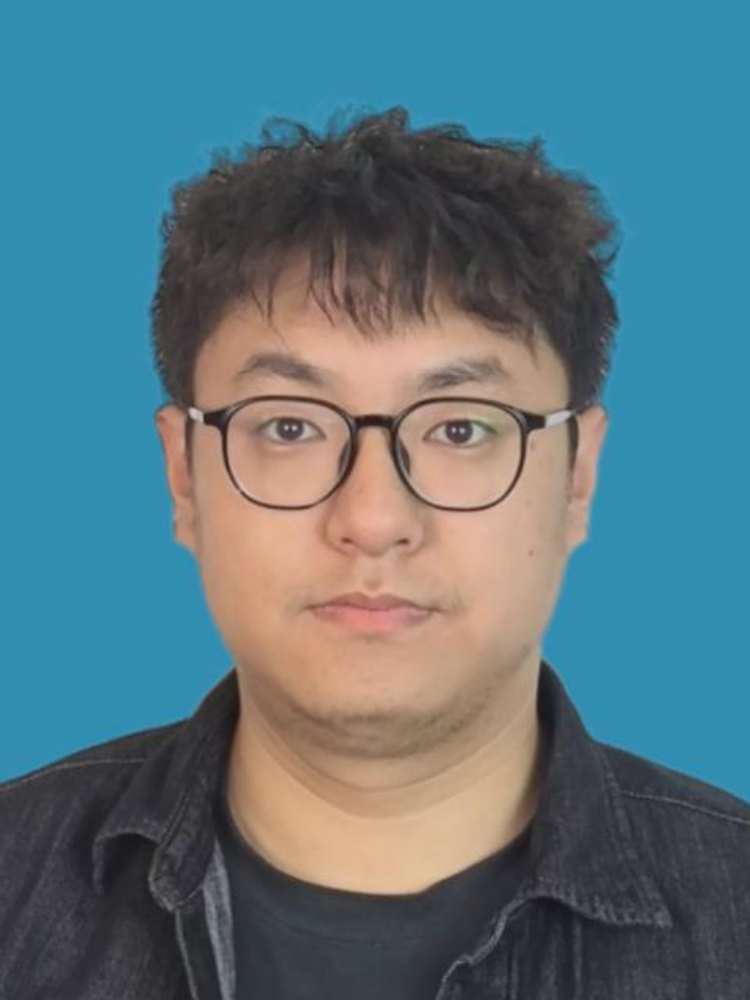}}]{Chenglong Huang} received the BS degree in communication engineering from Hohai University, China, in 2023. He is a graduate student from the School of Electronic Science and Engineering, Nanjing University. His research interests include spectral light-field imaging and deep optics.
\end{IEEEbiography}
\vspace{11pt}

\begin{IEEEbiography}[{\includegraphics[width=1in,height=1.25in,clip,keepaspectratio]{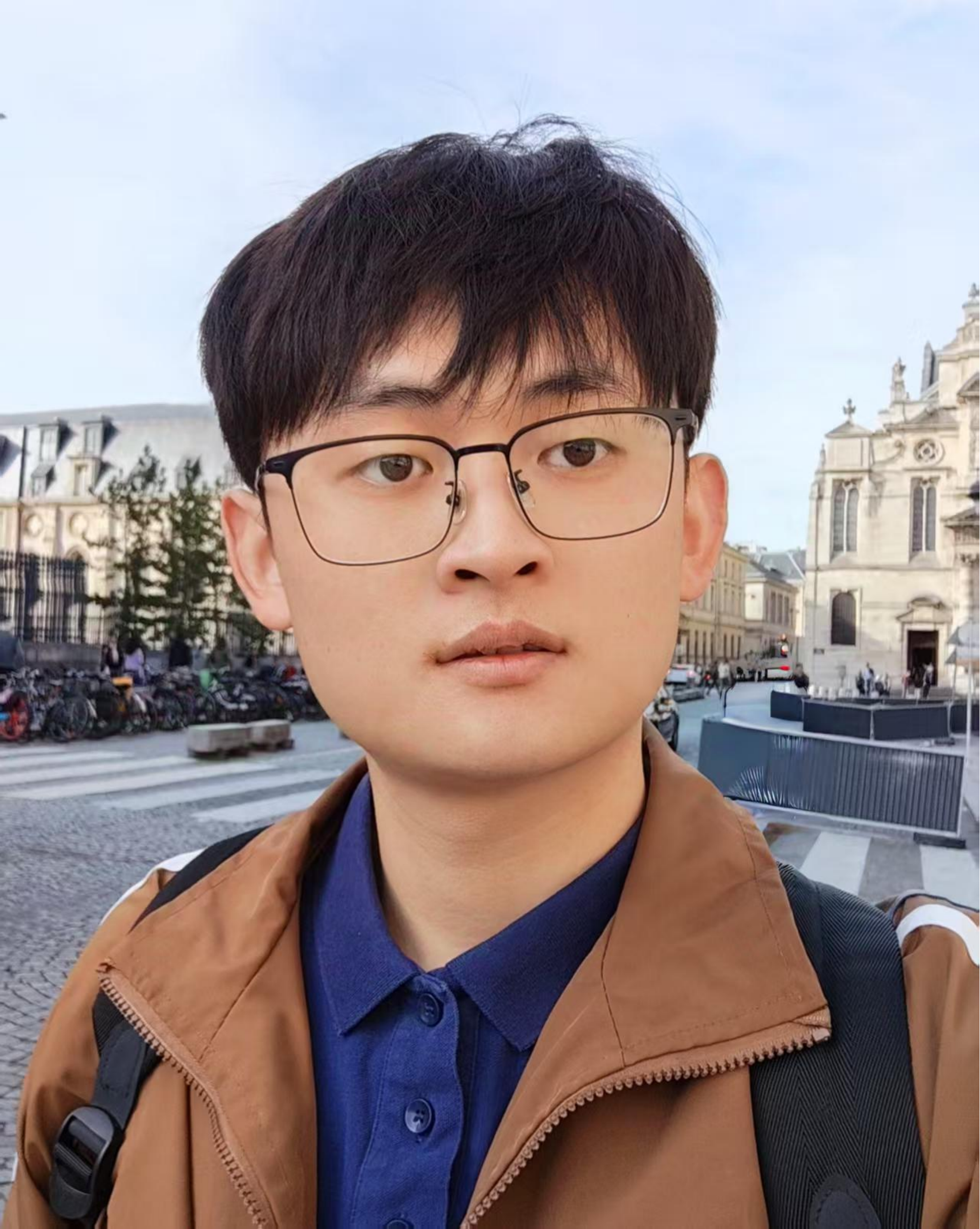}}]{Tao Lv} received the BS degree from the College of Information Science and Engineering, Northeastern University, Liaoning, China, in 2021. He is currently working toward the PhD degree with the School of Electronic Science and Engineering, Nanjing University, Nanjing, China. His research interests include computational photography and computer vision, especially computational spectral imaging.
\end{IEEEbiography}
\vspace{11pt}

\begin{IEEEbiography}[{\includegraphics[width=1in,height=1.25in,clip,keepaspectratio]{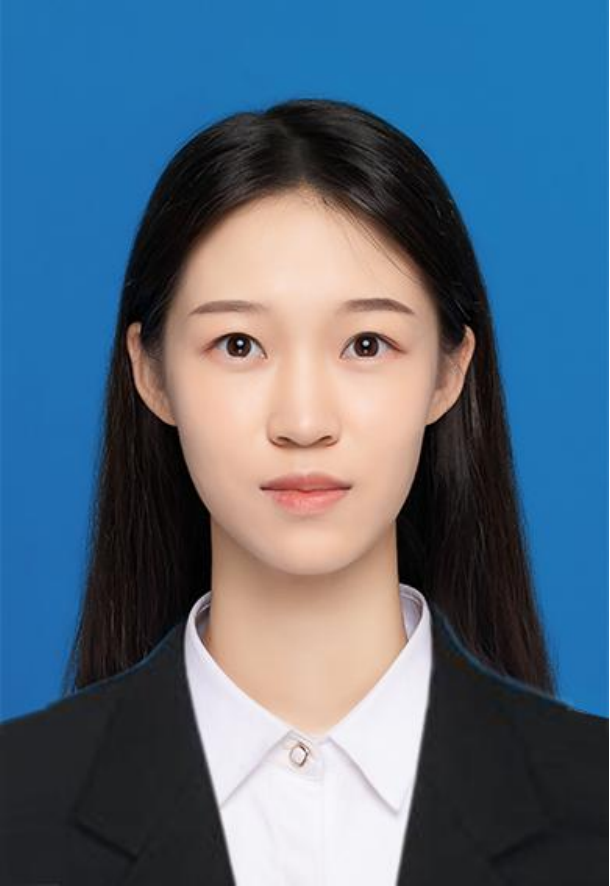}}]{Jianing Yang} received the BS degree in automation from Shanghai University Nanjing,China, in 2025. She is currently working toward the PhD degree with the School of Electronic Science and Engineering, Nanjing University. Her research interests include computational imaging and computer vision.
\end{IEEEbiography}
\vspace{11pt}



\begin{IEEEbiography}[{\includegraphics[width=1in,height=1.25in,clip,keepaspectratio]{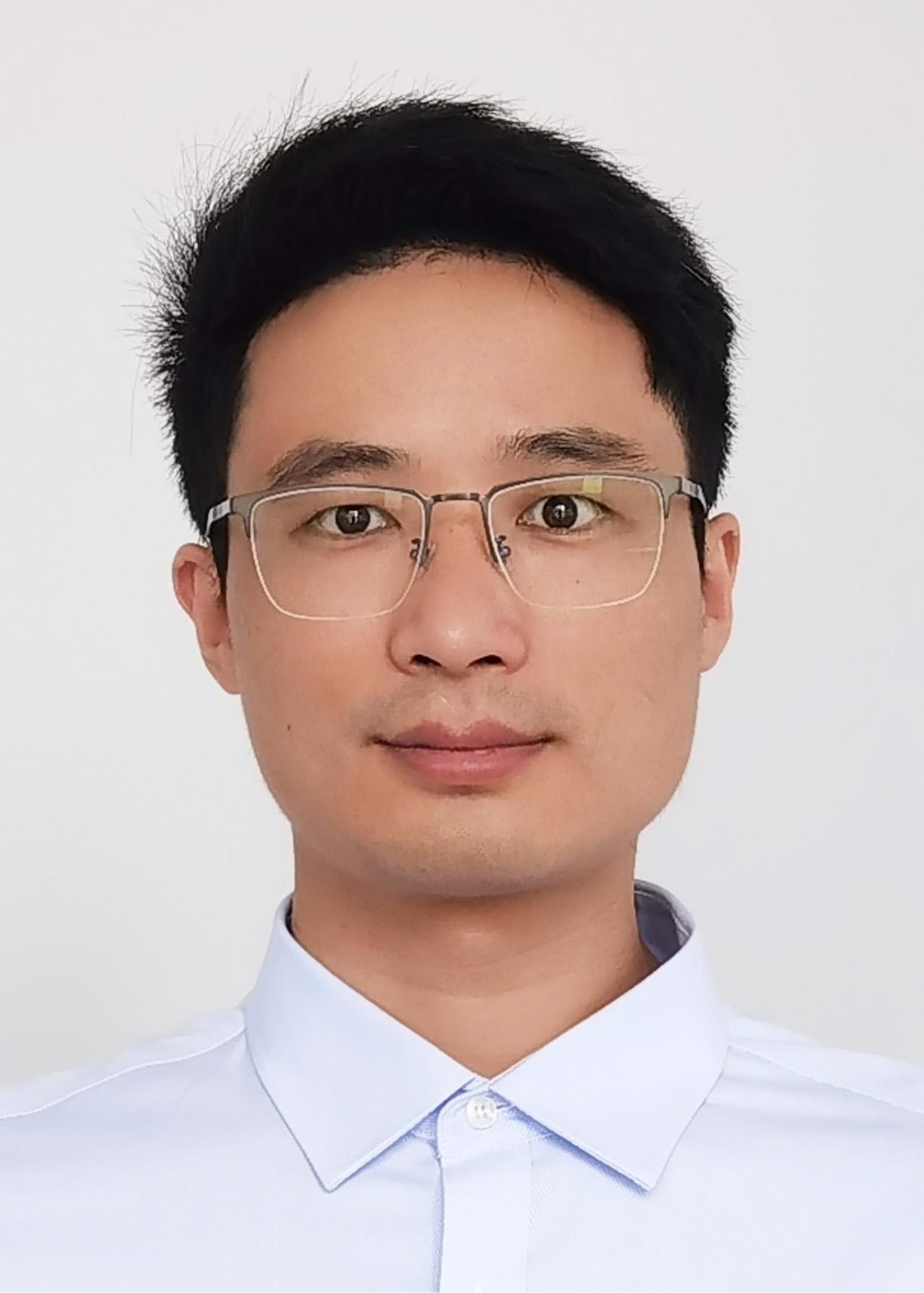}}]{Chongde Zi} received BS and MS degrees from Yunnan Normal University, Kunming, China, in 2014 and 2017, respectively, and PhD degree from Nanjing University, Nanjing, China, in 2024. He is currently a assistant researcher at Nanjing University, Nanjing, China. His research interests include computational photography and spectral imaging.
\end{IEEEbiography}
\vspace{11pt}

\begin{IEEEbiography}[{\includegraphics[width=1in,height=1.25in,clip,keepaspectratio]{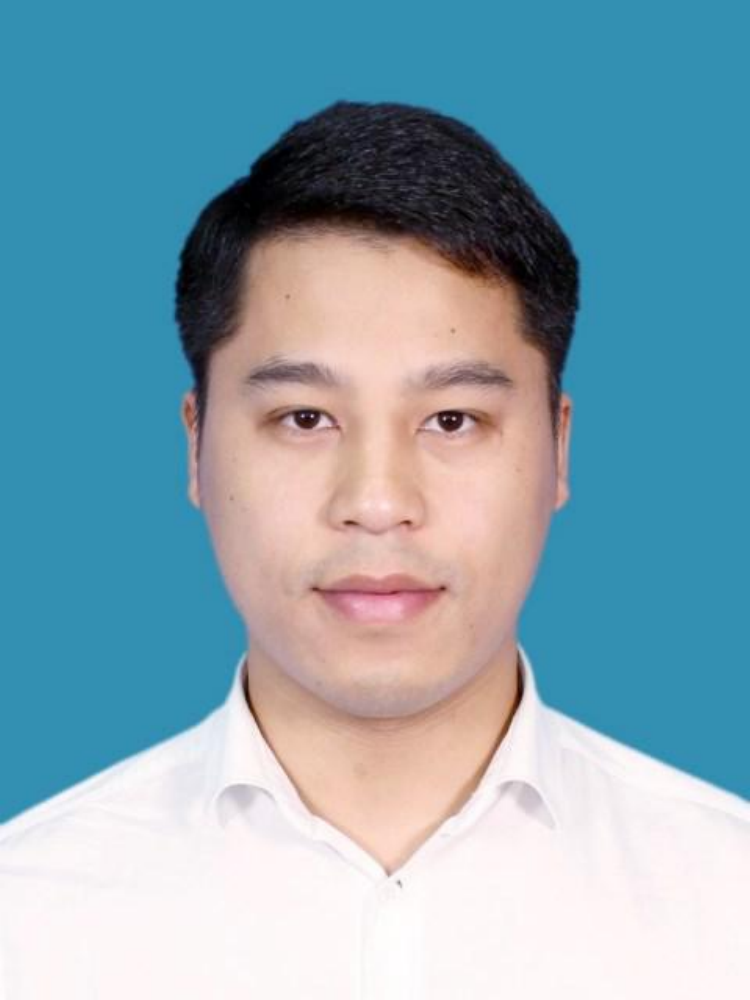}}]{Linsen Chen}
received the BS, MS and PhD degrees in 2014, 2017 and 2023 from School of Electronic Science and Engineering, Nanjing University, Nanjing, China. His research interests include computational photography, spectral imaging and reconstruction.
\end{IEEEbiography}
\vspace{11pt}


\begin{IEEEbiography}[{\includegraphics[width=1in,height=1.25in,clip,keepaspectratio]{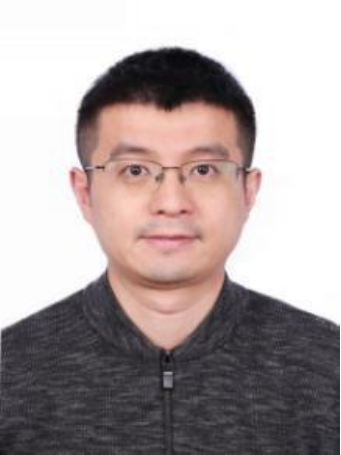}}]{Xun Cao}
(Member, IEEE) received the BS degree from Nanjing University, Nanjing, China, in 2006, and the PhD degree from the Department of Automation, Tsinghua University, Beijing, China, in 2012. He held visiting positions with Philips Research, Aachen, Germany, in 2008, and Microsoft Research Asia, Beijing, from 2009 to 2010. He was a visiting scholar with The University of Texas at Austin, Austin, Texas, from 2010 to 2011. He is currently a professor with the School of Electronic Science and Engineering, Nanjing University. His research interests include computational photography, image-based modeling and rendering, and VR/AR systems.
\end{IEEEbiography}

\vspace{11pt}


\vfill

\end{document}